\documentclass{article}

\usepackage{microtype}
\usepackage{graphicx}
\usepackage{subfigure}
\usepackage{booktabs} % for professional tables

\usepackage{url}
\usepackage{hyperref}

\usepackage[accepted]{icml2022}

\usepackage{amsmath}
\usepackage{amssymb}
\usepackage{mathtools}
\usepackage{amsthm}

\usepackage[capitalize,noabbrev,nameinlink]{cleveref}

\theoremstyle{plain}
\newtheorem{theorem}{Theorem}[section]

\newtheorem{lemma}[theorem]{Lemma}

\theoremstyle{definition}
\newtheorem{definition}[theorem]{Definition}

\theoremstyle{remark}

\usepackage[textsize=tiny]{todonotes}

\renewcommand{\algorithmiccomment}[1]{\hfill $\triangleright$ #1}
\usepackage{xcolor}
\usepackage{graphicx}
\usepackage{csquotes}
\usepackage{graphicx}
\usepackage{amsmath}
\usepackage{mathtools}
\usepackage{amssymb}
\usepackage{amsfonts}
\usepackage{bm}
\usepackage{booktabs}
\usepackage{mleftright}
\usepackage{bbm}
\usepackage{multirow}
\usepackage{enumitem}
\usepackage{wrapfig,lipsum}
\usepackage{tikz}
\usepackage{circledsteps}
\pgfkeys{/csteps/inner color=white}
\pgfkeys{/csteps/outer color=white}
\pgfkeys{/csteps/fill color=black}
\pgfkeys{/csteps/inner ysep=3pt}
\pgfkeys{/csteps/inner xsep=3pt}

\newcommand\simiid{\stackrel{\text{iid}}{\sim}}
\newcommand{\Bernoulli}{\operatorname{Bern}}
\newcommand\Unif{\operatorname{Unif}}

\newcommand{\oneg}{\vcenter{\hbox{\includegraphics[scale=0.1]{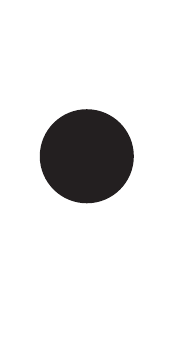}}}}
\newcommand{\twog}{\vcenter{\hbox{\includegraphics[scale=0.1]{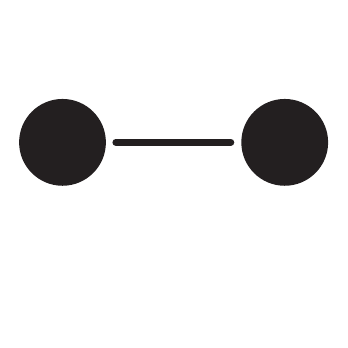}}}}
\newcommand{\threeg}{\vcenter{\hbox{\includegraphics[scale=0.1]{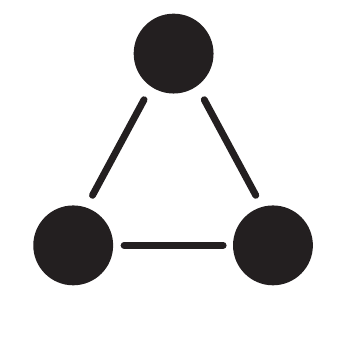}}}}
\newcommand{\sixg}{\vcenter{\hbox{\includegraphics[scale=0.2]{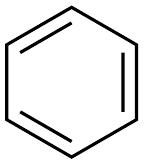}}}}

\newcommand{\reaarr}{\vcenter{\hbox{\includegraphics[scale=0.3]{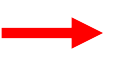}}}}
\newcommand{\pinkarr}{\vcenter{\hbox{\includegraphics[scale=0.3]{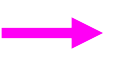}}}}
\newcommand{\pinkcircle}{\vcenter{\hbox{\includegraphics[scale=0.3]{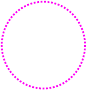}}}}

\newcommand{\ms}[2]{{#1\tiny{$\pm$#2}}}
\newcommand{\bms}[2]{{\textbf{#1}\tiny{$\pm$#2}}}

\icmltitlerunning{$\mathcal{G}$-Mixup: Graph Data Augmentation for Graph Classification}

\begin{document}

\twocolumn[
\icmltitle{$\mathcal{G}$-Mixup: Graph Data Augmentation for Graph Classification}



\icmlsetsymbol{equal}{*}

\begin{icmlauthorlist}
\icmlauthor{Xiaotian Han}{tamu}
\icmlauthor{Zhimeng Jiang}{tamu}
\icmlauthor{Ninghao Liu}{uga}
\icmlauthor{Xia Hu}{rice}
\end{icmlauthorlist}

\icmlaffiliation{tamu}{Department of Computer Science\&Engineering, Texas A\&M University}
\icmlaffiliation{uga}{Department of Computer Science, University of Georgia}
\icmlaffiliation{rice}{Department of Computer Science, Rice University}

\icmlcorrespondingauthor{Xiaotian Han}{han@tamu.edu}

\icmlkeywords{Machine Learning, ICML}

\vskip 0.3in
]



\printAffiliationsAndNotice{\icmlEqualContribution} % otherwise use the standard text.

\begin{abstract}
This work develops \emph{mixup for graph data}. Mixup has shown superiority in improving the generalization and robustness of neural networks by interpolating features and labels between two random samples. Traditionally, Mixup can work on regular, grid-like, and Euclidean data such as image or tabular data. However, it is challenging to directly adopt Mixup to augment graph data because different graphs typically: 1) have different numbers of nodes; 2) are not readily aligned; and 3) have unique typologies in non-Euclidean space. To this end, we propose $\mathcal{G}$-Mixup to augment graphs for graph classification by interpolating the generator (i.e., graphon) of different classes of graphs. Specifically, we first use graphs within the same class to estimate a graphon. Then, instead of directly manipulating graphs, we interpolate graphons of different classes in the Euclidean space to get mixed graphons, where the synthetic graphs are generated through sampling based on the mixed graphons. Extensive experiments show that $\mathcal{G}$-Mixup substantially improves the generalization and robustness of GNNs.
\end{abstract}

\section{Introduction}\label{sec:intro}
Recently deep learning has been widely adopted to graph analysis. Graph Neural Networks (GNNs)~\citep{wu2020comprehensive, zhou2020graph, zhang2020deep, xu2018powerful} have shown promising performance on graph classification. Meanwhile, data augmentation (e.g., DropEdge~\citep{rong2020dropedge}, Subgraph~\citep{you2020graph,wang2020graphcrop} ) has also been adopted to graph analysis by generating synthetic graphs to create more training data for improving the generalization of graph classification models. However, existing graph data augmentation strategies typically aim to augment graphs at a \emph{within-graph} level by either modifying edges or nodes in an individual graph, which does not enable information exchange between different instances. The \emph{between-graph} augmentation methods~(i.e., data augmentation between graphs) are still under-explored. 

In parallel with the development of graph neural networks, Mixup~\citep{zhang2017mixup} and its variants (e.g., Manifold Mixup~\citep{verma2019manifold}), as data augmentation methods, have been theoretically and empirically shown to improve the generalization and robustness of deep neural networks in image recognition~\citep{zhang2017mixup,verma2019manifold,zhang2021does} and natural language processing~\citep{guo2019augmenting, guo2020nonlinear}. The basic idea of Mixup is to linearly interpolate continuous values of random sample pairs to generate more synthetic training data. The formal mathematical expression of Mixup is
$\mathbf{x}_{new} = \lambda \mathbf{x}_i + (1-\lambda) \mathbf{x}_j, \mathbf{y}_{new} = \lambda \mathbf{y}_i + (1-\lambda) \mathbf{y}_j,$
where $(\mathbf{x}_i,\mathbf{y}_i )$ and $(\mathbf{x}_j,\mathbf{y}_j)$ are two samples drawn at random from training data and the target $\mathbf{y}$ are one-hot labels. With \emph{graph neural networks} and \emph{mixup} in mind, the following question naturally arises:
\begin{center}
    \vspace{-5pt}
    \textit{\textbf{Can we mix up graph data to improve the generalization and robustness of GNNs? } }
    \vspace{-5pt}
\end{center}
It remains an open and challenging problem to mix up graph data due to the characteristics of graphs and the requirements of applying Mixup. Typically, Mixup requires that original data instances are regular and well-aligned in Euclidean space, such as image data and table data. However, graph data is distinctly different from them due to the following reasons: \emph{(i) graph data is irregular}, since the number of nodes in different graphs are typically different from each other; \emph{(ii) graph data is not well-aligned}, where nodes in graphs are not naturally ordered and it is hard to match up nodes between different graphs; \emph{(iii) graph topology between classes are divergent}, where the topologies of a pair of graphs from different classes are usually different while the topologies of those from the same class are usually similar. Thus, it is nontrivial to directly adopt the Mixup strategy to graph data.

\begin{figure*}[!tb]
      \centering
      \includegraphics[width=1.0\textwidth]{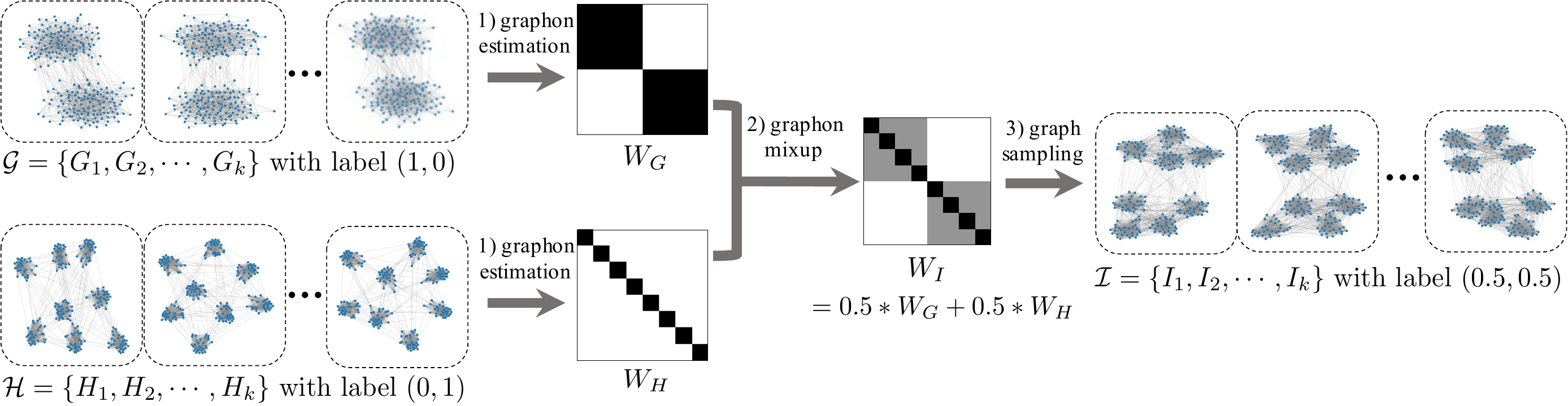}
      \vspace{-15pt}
      \caption{An overview of $\mathcal{G}$-Mixup. The task is binary graph classification. We have two classes of graphs $\mathcal{G}$ and $\mathcal{H}$ with different topologies ($\mathcal{G}$ has two communities while $\mathcal{H}$ has eight communities). $\mathcal{G}$ and $\mathcal{H}$ have different graphons. We mix up the graphons ${W}_{\mathcal{G}}$ and ${W}_{\mathcal{H}}$ to obtain a mixed graphon ${W}_{\mathcal{I}}$, and then sample new graphs from the mixed graphon. Intuitively, the synthetic graphs have two major communities and each of which has four sub-communities, demonstrating that the generated graphs preserve the structure of original graphs from both classes.}\label{fig:mixup:meth}
      \vspace{-5pt}
\end{figure*}

To tackle the aforementioned problems, we propose $\mathcal{G}$-Mixup, a class-level graph data augmentation method, to mix up graph data based on graphons. The graphs within one class are produced under the same generator (i.e., graphon). We mix up the graphons of different classes and then generate synthetic graphs. Informally, a graphon can be thought of as a probability matrix (e.g., the matrix $W_G$ and $W_H$ in \cref{fig:mixup:meth}), where $W(i,j)$ represents the probability of edge between node $i$ and $j$. The real-world graphs can be regraded as generated from graphons. Since the graphons of different graphs is regular, well-aligned, and is defined in Euclidean space, it is easy and natural to mix up the graphons and then generate the synthetic graphs therefrom. On this basis, we can achieve graphs mixup by mixing their generators. We also provide theoretical analysis of graphons mixup, which guarantees that the generated graphs will preserve the key characteristics of both original classes. Our proposed method is illustrated in \cref{fig:mixup:meth} with an example. Given two graph training sets $\mathcal{G} = \{G_{1}, G_{2}, \cdots, G_{m}\}$ and $\mathcal{H}= \{H_{1}, H_{2}, \cdots, H_{m}\}$ with different labels and distinct topologies (i.e., $\mathcal{G}$ has two communities while $\mathcal{H}$ has eight communities), we estimate graphons ${W}_{\mathcal{G}}$ and ${W}_{\mathcal{H}}$ respectively from $\mathcal{G}$ and $\mathcal{H}$. We then mix up the two graphons and obtain a mixed graphon ${W}_{\mathcal{I}}$. After that, we sample synthetic graphs from ${W}_{\mathcal{I}}$ as additional training graphs. The generated synthetic graphs have two major communities and each of them have four sub-communities, which is a mixture of the two graph sets. It thus shows that $\mathcal{G}$-Mixup is capable of mixing up graphs.

Our \textbf{main contributions} are highlighted as follows: \textit{Firstly}, we propose $\mathcal{G}$-Mixup to augment the training graphs for graph classification. Since directly mixing up graphs is intractable, $\mathcal{G}$-Mixup mixes the graphons of different classes of graphs to generate synthetic graphs. 
\textit{Secondly}, we theoretically prove that the synthetic graph will be the mixture of the original graphs, where the key topology (i.e., discriminative motif) of source graphs will be mixed up.
\textit{Thirdly}, we demonstrate the effectiveness of the proposed $\mathcal{G}$-Mixup on various graph neural networks and datasets. Extensive experimental results show that $\mathcal{G}$-Mixup substantially improves the performance of graph neural networks in terms of enhancing their generalization and robustness.

\section{Preliminaries}\label{sec:back}

In this section, we first go over the notations used in this paper, and then introduce graph related concepts including graph homomorphism and graphons, which will be used for theoretical analysis in this work. Finally, we briefly review the graph neural networks for graph classification.

\subsection{Notations}
Given a graph $G$, we use $V(G)$ and $E(G)$ to denote its nodes and edges, respectively. The number of nodes is $\mathrm{v}(G)=|V(G)|$, and the number of edges is $\mathrm{e}(G)=|E(G)|$. We use $m,l$ to denote the number of graphs and $N,K$ to denote the number of nodes. We use $G,H,I$/$\mathcal{G,H,I}$ to denote graphs/graph set. $\mathbf{y}_\mathcal{G} \in \mathbb{R}^{C}$ denotes the label of graph set $\mathcal{G}$, where $C$ is number of classes of graphs. A graph could contain some frequent subgraphs which are called \textit{motifs}. The motifs in graph $G$ is denoted as $F_{G}$. The set of motifs in graph set $\mathcal{G}$ is denoted as $\mathcal{F}_{\mathcal{G}}$. 
 $W_\mathcal{G}$ denotes the graphon of graph set $\mathcal{G}$. $\mathbf{W}$ denotes the step function. $\mathbb{G}(K,W)$ denotes the random graph with $K$ nodes based on graphon $W$.

\subsection{Graph Homomorphism and Graphons}

\textbf{Graph Homomorphism.} A graph homomorphism is an adjacency-preserving mapping between two graphs, i.e., mapping adjacent vertices in one graph to adjacent vertices in the other. Formally, a \emph{graph homomorphism} $\phi \colon F\to G$ is a map from $V(F)$ to $V(G)$, where if $\{u,v\} \in E(F)$, then $\{\phi(u),\phi(v)\} \in E(G)$.
For two graphs $H$ and $G$, there could be multiple graph homomorphisms between them. Let $\hom (H,G)$ denotes the total number of graph homomorphisms from graph $H$ to graph $G$. For example, $\hom(\oneg,G) = |V(G)|$ if graph $H$ is $\oneg$, $\hom(\twog,G) = 2 |E(G)|$ if graph $H$ is $\twog$, and $\hom(\threeg,G)$ is six times the number of triangles in $G$. There are in total $|V(G)|^{|V(H)|}$ mappings from $H$ to $G$, but only some of them are homomorphisms. Thus, we define \emph{homomorphism density} to measure the relative frequency that the graph $H$ appears in graph $G$ as $t(H,G) = \frac{\hom(H,G)}{|V(G)|^{|V(H)|}}$. For example, $t(\oneg,G) = |V(G)| / N^1=1$, $t(\twog,G) = 2 |E(G)| / N^2$.

\textbf{Graphon.} A graphon~\citep{airoldi2013stochastic} is a continuous, bounded and symmetric function $W: [0,1]^2 \to [0,1]$ which may be thought of as the weight matrix of a graph with infinite number of nodes. Then, given two points $u_i$, $u_j \in [0,1]$, $W(i,j)$ represents the probability that nodes $i$ and $j$ be related with an edge. 
Various quantities of a graph can be calculated as a function of the graphon. For example, the degree of nodes in graphs can be easily extended to a degree distribution function in graphons, which is characterized by its graphon marginal $d_W(x) = \int_{0}^{1}W(x,y)dy$. Similarly, the concept of homomorphism density can be naturally extended from graphs to graphons. Given an arbitrary graph motif $F$, its homomorphism density with respect to graphon $W$ is defined by $t(F,W) = \int_{[0,1]^{V(F)}}\prod_{i, j\in E(F)}W(x_i,x_j) \prod_{i\in V(F)}dx_i$.
For example, the edge density of graphon $W$ is $t(\twog,W) = \int_{[0,1]^2} W(x,y) \ dx dy$, and the triangle density of graphon $W$ is $t(\threeg,W) = \int_{[0,1]^3} W(x,y)W(x,z)W(y,z) \ dx dy dz$.

\subsection{Graph Classification with Graph Neural Networks}
Given a set of graphs, graph classification aims to assign a class label for each graph $G$. Recently, graph neural networks have become the state-of-the-art approach for graph classification. Without loss of generalization, we present the formal expression of a graph convolution network~(GCN)~\citep{kipf2016semi}. The forward propagation at $k$-th layer is described as the following:
\begin{equation}\label{eq:combine}
    \begin{aligned}
        &\mathbf{a}_i^{(k)} = \text{AGG}^{(k)} \left( \left\lbrace \mathbf{h}_j^{(k-1)} : j \in \mathcal{N}(i) \right\rbrace \right), \\
        &\mathbf{h}_i^{(k)} = \text{COMBINE}^{(k)} \left( \mathbf{h}_i^{(k-1)}, \mathbf{a}_i^{(k)} \right),
    \end{aligned}
\end{equation}
\noindent where $\mathbf{h}_i^{(k)}\in\mathbb{R}^{n\times d_k}$ is the intermediate representation of node $i$ at the $k$-th layer, $\mathcal{N}(i)$ denotes the neighbors of node $i$. $\mathrm{AGG}(\cdot)$ is an aggregation function to collect embedding representations from neighbors, and $\mathrm{COMBINE}(\cdot)$ combines neighbors' representation and its representation at $(k-1)$-th layer. For graph classification, a graph-level representation is obtained by summarizing all node-level representations in the graph by a readout function:
\begin{equation}\label{eq:readout}
    \begin{aligned}
        &\mathbf{h}_{G} = \mathrm{READOUT} \left( \left\lbrace \mathbf{h}_i^{ (k) } : i \in E(G) \right\rbrace \right), \\
        &\hat{\mathbf{y}} = \mathrm{softmax}( \mathbf{h}_{G} ),
    \end{aligned}
\end{equation}
where $\mathrm{ READOUT}(\cdot)$ is the readout function, which can be a simple function such as average or sophisticated pooling function~\citep{gao2019graph,ying2018hierarchical}, $\mathbf{h}_{G}$ is the representation of graph $G$, and $\hat{\mathbf{y}}\in \mathbb{R}^C$ is the predicted probability that $G$ belongs to each of the $C$ classes.

\section{Methodology}\label{sec:meth}

In this section, we formally introduce the proposed $\mathcal{G}$-Mixup and elaborate its implementation details. 

\subsection{\texorpdfstring{ $\mathcal{G}$ }~-Mixup}\label{sec:meth:mixu}
Different from the interpolation of image data in Euclidean space, adopting Mixup to graph data is nontrivial since graphs are irregular, unaligned and non-Euclidean data. In this work, we show that the challenges could be tackled with graphon theory. Intuitively, a graphon can be regarded as a graph generator. Graphs of the same class can be seen as being generated from the same graphon. With this in mind, we propose $\mathcal{G}$-Mixup, a class-level data augmentation method via graphon interpolation. Specifically, $\mathcal{G}$-Mixup interpolates different graph generators to obtain a new mixed one. Then, synthetic graphs are sampled based on the mixed graphon for data augmentation. The graphs sampled from this generator partially possess properties of the original graphs. Formally, $\mathcal{G}$-Mixup is formulated as:
\begin{alignat}{2}
% \begin{split}
   &\text{Graphon Estimation:}&\mathcal{G} \to {W}_{\mathcal{G}}, \mathcal{H} \to {W}_{\mathcal{H}}\label{equ:gmixup:gone}\\
   &\text{Graphon Mixup:}&{W}_{\mathcal{I}} = \lambda {W}_{\mathcal{G}} + (1-\lambda) {W}_{\mathcal{H}}\label{equ:gmixup:mixw}\\
   &\text{Graph Generation:}&\{{I}_1, {I}_2, \cdots ,{I}_m\}  \stackrel{\text{i.i.d}}{\sim} \mathbb{G}(K, {W}_{\mathcal{I}})\label{equ:gmixup:resam}\\
   &\text{Label Mixup:}&\mathbf{y}_{ \mathcal{I} } = \lambda \mathbf{ y }_{ \mathcal{G} } + (1-\lambda) \mathbf{y}_{\mathcal{H}}\label{equ:gmixup:mixy}
% \end{split}
\end{alignat}
where ${W}_{\mathcal{G}}, {W}_{\mathcal{H}}$ are graphons of the graph set $\mathcal{G}$ and $\mathcal{H}$. The mixed graphon is denoted by $ {W}_{\mathcal{I}}$, and $\lambda \in [0,1]$ is the trade-off hyperparameter to control the contributions from different source sets. The set of synthetic graphs generated from $ {W}_{\mathcal{I}}$ is $\mathcal{I} = \{{I}_1, {I}_2, \cdots ,{I}_m\}$. The $\mathbf{y}_{ \mathcal{G} }\in \mathbb{R}^C$ and $\mathbf{y}_{\mathcal{H}}\in \mathbb{R}^C$ are vectors containing ground-truth labels for graph $G$ and $H$, respectively, where $C$ is the number of classes. The label vector of synthetic graphs in graph set $\mathcal{I}$ is denoted as $\mathbf{y}_{ \mathcal{I} }\in \mathbb{R}^C$.

As illustrated in \cref{fig:mixup:meth} and the above equations, the proposed $\mathcal{G}$-Mixup includes three key steps: \textbf{i) estimate a graphon} for each class of graphs, \textbf{ii) mix up the graphons} of different graph classes, and \textbf{iii) generate synthetic graphs} based on the mixed graphons. Specifically, suppose we have two graph sets $\mathcal{G} = \{ {G}_1,{G}_2,\cdots,{G}_m \} $ with label $\mathbf{y}_{ {G} }$, and $\mathcal{H} = \{{H}_1,{H}_2,\cdots,{H}_m \}$ with label $\mathbf{y}_{ \mathcal{H} }$. Graphons ${W}_{ \mathcal{G} }$ and ${W}_{ \mathcal{H} }$ are estimated from graph sets $\mathcal{G}$ and $\mathcal{H}$, respectively. Then, we mix them up by linearly interpolating the two graphons and their labels, and obtain ${W}_{\mathcal{I}}$ and $\mathbf{y}_{\mathcal{I}}$. Finally, a set of synthetic graphs $\mathcal{I}$ is sampled based on ${W}_{\mathcal{I}}$, which will be used as additional training graph data.

\subsection{Implementation}\label{sec:meth:impl}

In this section, we introduce the implementation details of graphon estimation and synthetic graphs generation. We provide the pseudo-code of $\mathcal{G}$-Mixup in \cref{sec:appe:imp_detail}.

\noindent\textbf{Graphon Estimation and Mixup.}
Estimating graphons from observed graphs is a prerequisite for $\mathcal{G}$-Mixup.
However, it is intractable because a graphon is an unknown function without a closed-form expression for real-world graphs. Therefore, we use the step function~\citep{lovasz2012large, xu2021learning} to approximate graphons\footnote{Because weak regularity lemma of graphon~\citep{frieze1999quick} indicates that an arbitrary graphon can be approximated well by step function. Detailed discussion is in \cref{sec:appe:step}.}. In general, the \emph{step function} can be seen as a matrix $\mathbf{W} = [w_{kk'}] \in [0,1]^{K\times K}$, where $\mathbf{W}_{ij}$ is the probability that an edge exists between node $i$ and node $j$. In practice, we use the matrix-form step function as graphon to mix up and generate synthetic graphs. The step function estimation methods are well-studied, which first align the nodes in a set of graphs based on node measurements (e.g., degree) and then estimate the step function from all the aligned adjacency matrices. The typical step function estimation methods includes sorting-and-smoothing (SAS) method~\citep{chan2014consistent}, stochastic block approximation (SBA)~\citep{airoldi2013stochastic}, ``largest gap'' (LG)~\citep{channarond2012classification}, matrix completion (MC)~\citep{keshavan2010matrix}, universal singular value thresholding (USVT)~\citep{chatterjee2015matrix}.~\footnote{The details about these step function estimation methods are presented in~\cref{sec:appe:gonm}.}
Formally, a step function $\mathbf{W}^{P}:[0,1]^2\mapsto [0, 1]$ is defined as $\mathbf{W}^{P}(x,y)=\sideset{}{_{k,k'=1}^{K}}\sum w_{kk'} \mathbbm{1}_{\mathcal{P}_k\times \mathcal{P}_{k'}}(x,y)$,
where $\mathcal{P}=(\mathcal{P}_1,..,\mathcal{P}_K)$ denotes the partition of $[0, 1]$ into $K$ adjacent intervals of length $1/K$, $w_{kk'}\in [0, 1]$, and indicator function $\mathbbm{1}_{\mathcal{P}_k\times \mathcal{P}_{k'}}(x,y)$ equals to 1 if $(x, y)\in\mathcal{P}_{k}\times\mathcal{P}_{k'}$ and otherwise it is $0$.
\emph{For binary classification}, we have $\mathcal{G} = \{ G_1, G_2,\cdots,G_m\}$ and $\mathcal{H} = \{{H}_1,{H}_2,\cdots,{H}_m \}$ with different labels, we estimate their step functions $\mathbf{W}_\mathcal{G}\in \mathbb{R}^{K\times K}$ and $\mathbf{W}_\mathcal{H}\in \mathbb{R}^{K\times K}$,
where we let $K$ be the average number of nodes in all graphs. \emph{For multi-class classification}, we first estimate the step function for each class of graphs and then randomly select two to perform mix-up. 
The resultant step function is $\mathbf{W}_\mathcal{I} = \lambda \mathbf{W}_\mathcal{G} + (1-\lambda)\mathbf{W}_\mathcal{H} \in \mathbb{R}^{K\times K}$, which serves as the generator of synthetic graphs.

\textbf{Synthetic Graphs Generation.} A graphon $W$ provides a distribution to generate arbitrarily sized graphs. Specifically, a $K$-node random graph $\mathbb{G}(K, {W}_{\mathcal{I}})$ can be generated following the process:
\begin{align}
       u_1, \dots, u_K \simiid \Unif_{[0,1]},\ &\mathbb{G}(K, W)_{ij} \simiid \Bernoulli(W(u_i,u_j)),\nonumber \\
       &\forall i,j \in [K]. \nonumber
\end{align}
Since we use the step function $\mathbf{W}$ to approximate the graphon $W$, we set $W(u_i,u_j) = \mathbf{W}[\lfloor 1/u_i \rfloor, \lfloor 1/u_j \rfloor]$, and $\lfloor \cdot \rfloor$ is the floor function. The first step samples $K$ nodes independently from a uniform distribution $\Unif_{[0,1]}$ on $[0,1]$. The second step generates an adjacency matrix $\bm{A}=[a_{ij}]\in \{0, 1\}^{K\times K}$, whose element values follow the Bernoulli distributions $\Bernoulli( \cdot )$  determined by the 
step function. A graph is thus obtained as $G$ where $V(G)=\{1,...,K\}$ and $E(G)=\{(i,j)~|~a_{ij}=1\}$. A set of synthetic graphs can be generated by conducting the above process multiple times. The generation of node features of synthetic graphs includes two steps: 1) build the graphon node features based on the original node features, 2) generate node features of synthetic graphs based on the graphon node features. Specifically, at the graphon estimation phase, we align original node features while aligning the adjacency matrices, so we have a set of aligned original node features for each graphon, then we conduct pooling (average pooling in our experiments) on the aligned original node features to obtain the \emph{graphon node features}. The node features of generated graphs are the same as graphon features.

\begin{wraptable}{r}{0.45\linewidth}
\vspace{-10pt}
% \begin{table}[t]
    \centering  
    \fontsize{5}{8}\selectfont
    \setlength{\tabcolsep}{2pt}
    \caption{Computational complexity of graphon estimation.}\label{tab:complexity}
    % \vspace{-10pt}
    \begin{tabular}{lr}
    \toprule
    Method & Complexity \\
    \midrule
    MC   &$\mathcal{O}(N^3)$  \\
    USVT &$\mathcal{O}(N^3)$  \\
    LG   &$\mathcal{O}(mN^2)$ \\
    SBA  &$\mathcal{O}(mKN\log N)$ \\
    SAS  &$\mathcal{O}(mN\log N + K^2\log K^2)$\\
    \bottomrule
    \end{tabular}
% \end{table}
\vspace{-15pt}
\end{wraptable}

\noindent\textbf{Computational Complexity Analysis.} We hereby discuss computational complexity of $\mathcal{G}$-Mixup. The major computation costs come from graphon estimation and synthetic graph generation. 
For graphon estimation, suppose we have $m$ graphs and each of them has $N$ nodes, and estimate step function with $K$ partitions to approximate a graphon, the complexity of used graphon estimation methods~\citep{xu2021learning} is in \cref{tab:complexity}. For graph generation, suppose we need to generate $l$ graphs with $K$ nodes, the computational complexity is $\mathcal{O}(lK)$ for node generation and $\mathcal{O}(lK^2)$ for edge generation, so the overall complexity of graph generation is $\mathcal{O}(lK^2)$.

\section{Theoretical Justification}\label{sec:meth:theo}

In the following, we theoretically prove that: \emph{the synthetic graphs generated by $\mathcal{G}$-Mixup will be a mixture of original graphs}. We first define the discriminative motif, and then we justify the graphon mixup operation (\cref{equ:gmixup:mixw}) and graph generation operation (\cref{equ:gmixup:resam}) by analysing the homomorphism density of discriminative motifs of the original graphs and the synthetic graphs.

\begin{definition}[Discriminative Motif]
A discriminative motif $F_{G}$ of graph $G$ is the subgraph, with the minimal number of nodes and edges, that can decide the class the graph $G$. Furthermore, $\mathcal{F}_{ \mathcal{G} }$ is the set of discriminative motifs for graphs in the set $\mathcal{G}$.
\end{definition}

Intuitively, the discriminative motif is the key topology of a graph. We assume that \emph{ (i) every graph $G$ has a discriminative motif $F_{G}$ }, and  \emph{(ii) a set of graphs $\mathcal{G}$ has a finite set of discriminative motifs $\mathcal{F}_{\mathcal{G}}$.} The goal of graph classification is to filter out structural noise in graphs~\citep{fox2019robust} and recognize the key typologies (discriminative motifs) to predict the class label. For example, benzene (a chemical compound) is distinguished by the motif $\sixg$ (benzene ring). In the following, we analyze $\mathcal{G}$-Mixup from the perspective of discriminative motifs.

\subsection{Will discriminative motifs \texorpdfstring{$F_\mathcal{G}$}~ and \texorpdfstring{$F_\mathcal{H}$}~~exist in  \texorpdfstring{$\lambda {W}_{\mathcal{G}} + (1-\lambda) {W}_{\mathcal{H}} $}~ ?}\label{sec:theo:q1}

We answer this question by exploring the difference in homomorphism density of discriminative motifs between the original and mixed graphon, as the following theorems,
\begin{theorem}\label{theo:bound}
Given two sets of graphs $\mathcal{G}$ and $\mathcal{H}$, the corresponding graphons are ${W}_{\mathcal{G}}$ and ${W}_{\mathcal{H}}$, and the corresponding discriminative motif set $\mathcal{F}_\mathcal{G}$ and $\mathcal{F}_\mathcal{H}$. For every discriminative motif $F_\mathcal{G} \in \mathcal{F}_\mathcal{G} $ and $F_\mathcal{H}\in \mathcal{F}_\mathcal{H}$, the difference between the homomorphism density of $F_\mathcal{G}$/$F_\mathcal{H}$ in the mixed graphon ${W}_{\mathcal{I}} = \lambda {W}_{\mathcal{G}} + (1-\lambda) {W}_{\mathcal{H}}$ and that of the graphon ${W}_{\mathcal{H}}$/${W}_{\mathcal{G}}$ is upper bounded by
\vspace{-3pt}
\begin{equation}\label{equ:theo1}
\begin{aligned}
        \left|t(F_\mathcal{G}, {W}_{\mathcal{I}}  ) - t(F_\mathcal{G},{W}_{\mathcal{G}} ) \right| &\leq (1-\lambda) \mathrm{e}(F_\mathcal{G})|| {W}_{\mathcal{H}} - {W}_{\mathcal{G}} ||_{\square} ,\\
        |t(F_\mathcal{H}, {W}_{\mathcal{I}} ) - t(F_\mathcal{H},{W}_{\mathcal{H}} ) | &\leq \lambda \mathrm{e}(F_\mathcal{H})|| {W}_{\mathcal{H}} - {W}_{\mathcal{G}} ||_{\square} \nonumber
\end{aligned}
\end{equation}
\vspace{-2pt}
where $\mathrm{e}(F)$ is the number of nodes in graph $F$, and $||{W}_{\mathcal{H}} - {W}_{\mathcal{G}}||_{\square}$ denotes the cut norm~\footnote{Cut norm is used to measure the similarity between graphs, Details about cut norm are in \cref{sec:appe:prel}}.

\end{theorem}
\textit{Proof Sketch.} The proof follows the derivation of Counting Lemma for Graphons~(Lemma 10.23 in~\citet{lovasz2012large}), which associates the homomorphism density with the cut norm $||{W}_{\mathcal{H}} - W_\mathcal{G}||_{\square}$ of graphons. Specifically, we take the two graphons in this Lemma to deduce the bound of the difference of homomorphism densities of ${W}_{\mathcal{I}}$ and  ${W}_{\mathcal{G}}/{W}_{\mathcal{H}}$. Detailed proof are in \cref{sec:appe:proof_p1}. $\hfill\blacksquare$

\cref{theo:bound} suggests that the difference in the homomorphism densities of the mixed graphon and original graphons is upper bounded. Note that difference depends on the hyperparameter $\lambda$, the edge number $\mathrm{e}(F_\mathcal{G})/\mathrm{e}(F_\mathcal{H})$ and the cut norm $|| {W}_{\mathcal{H}} - {W}_{\mathcal{G}} ||_{\square} $. Since the $\mathrm{e}(F_\mathcal{G})/\mathrm{e}(F_\mathcal{H})$ and the cut norm $|| {W}_{\mathcal{H}} - {W}_{\mathcal{G}} ||_{\square} $ are decided by the dataset~(can be seen as a constant), the difference in homomorphism densities will be decided by $\lambda$. On this basis, the label of the mixed graphon is set to $\lambda \mathbf{ y }_{ \mathcal{G} } + (1-\lambda) \mathbf{y}_{\mathcal{H}}$.
\textbf{Therefore, $\mathcal{G}$-Mixup can preserve the different discriminative motifs of the two different graphons into one mixed graphon.}

\subsection{Will the generated graphs from graphon \texorpdfstring{${W}_{\mathcal{I}}$}~ preserve the mixture of discriminative motifs?}\label{sec:theo:q2}

Ideally, the generated graphs should inherit the homomorphism density of discriminative motifs from the graphon. To verify this, we propose the following theorem.
\begin{theorem}\label{theo:sample}
Let ${W}_{\mathcal{I}}$ be the mixed graphon, $n\geq 1$, $0<\varepsilon<1$, and let $F_{\mathcal{I}}$ be the mixed discriminative motif, then the ${W}_{\mathcal{I}}$-random graph $\mathbb{G}=\mathbb{G}(n, {W}_{\mathcal{I}})$ satisfies
\vspace{-8pt}
\begin{equation}\label{equ:theo2}
    \mathrm{P} \left( |t({F}_\mathcal{I}, \mathbb{G} )-t( {F}_{\mathcal{I}},{W}_{\mathcal{I}} )|>\varepsilon \right) \leq 2\mathrm{exp}\left(-\frac{\varepsilon^2n}{8 \mathrm{v}(F_I)^2 } \right). \nonumber
\end{equation}
\end{theorem}
\vspace{-8pt}

\cref{theo:sample}~states that for any specified nonzero margin $\varepsilon$, with a sufficient number of graphs sampled from the mixed graphon, the homomorphism density of discriminative motif in synthetic graphs will approximately equal to that in graphon $t({F}_\mathcal{I}, \mathbb{G} ) \approx t( {F}_{\mathcal{I}},{W}_{\mathcal{I}} )$ with high probability. In other words, the synthetic graphs will preserve the discriminative motif of the mixed graphon with a very high probability if the sample number $n$ is large enough. The detailed proof is in \cref{sec:appe:proof_c1}. \textbf{Therefore, $\mathcal{G}$-Mixup can preserve the discriminative motifs of the two different graphs into one mixed graph.}

\vspace{-5pt}
\begin{figure*}[ht]
      \centering
      \includegraphics[width=1.0\textwidth]{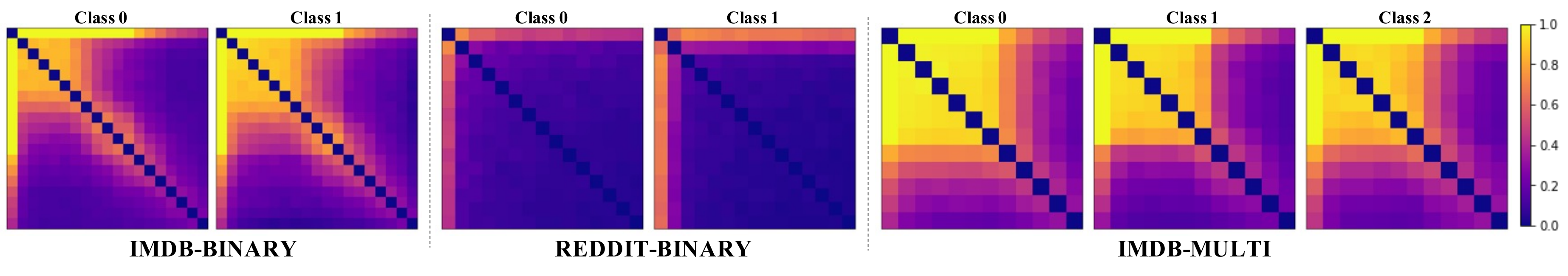}
      \vspace{-23pt}
      \caption{Estimated graphons on IMDB-BINARY, REDDIT-BINARY, and IMDB-MULTI. Obviously, graphons of different graph classes are quiet different. This observation validates the divergence of graphons between different classes of graphs, which is the basis of the $\mathcal{G}$-Mixup. The graphons are estimated by $LG$. More estimated graphons via various methods are in \cref{sec:appe:vis}. }\label{fig:gon}
\end{figure*}

\begin{figure*}[tb]
 \centering
 \includegraphics[width=1\textwidth]{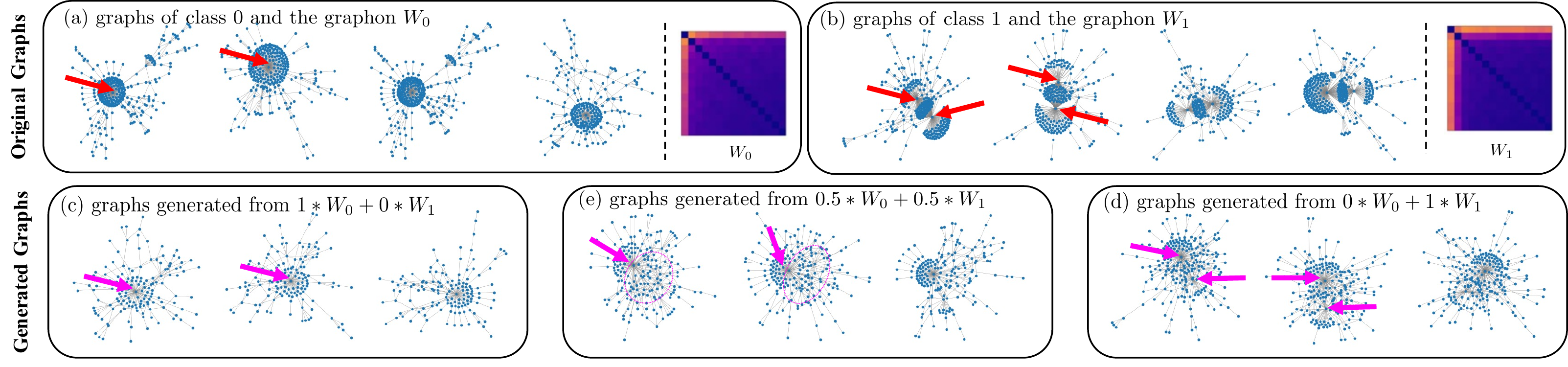}
  \vspace{-20pt} 
  \caption{The visualization of generated synthetic graphs on REDDIT-BINARY dataset. The first row is the original graphs while the second row is the generated graphs from $\mathcal{G}$-Mixup. The graphs in (a) and (b) are the original graphs of class 0 and class 1. The distinct difference between the two classes is that graphs of class 0 have one high-degree node while graphs of class 1 have two (marked with $\reaarr$in (a) and (b)). (c)/(d) shows graphs generated with the mixed graphon $(1*W_0 + 0*W_1)$ / $(0*W_0 + 1*W_1)$, which have one/two high-degree node/nodes (marked with $\pinkarr$in (c) and (d)) because the mixed graphon only contains $W_0$/$W_1$. The synthetic graphs generated from $(0.5*W_0 + 0.5*W_1)$ is the mixture of graphs of class $0$ and $1$, which appears as a high-degree node and a dense subgraph (marked with $\pinkarr$ and $\pinkcircle$ in (e), respectively). The results show that synthetic graphs are the mixture of the original graphs. }\label{fig:sampling}
 \vspace{-10pt} 
\end{figure*}

\subsection{Discussion}\label{sec:meth:disc}
We discuss the differences and relations between $\mathcal{G}$-Mixup and other augmentation strategies. 

\textbf{Relation to Edge Perturbation Methods}. The commonly used edge perturbation methods are spacial cases of $\mathcal{G}$-Mixup. Edge perturbation methods randomly perturb the edges to improve the GNNs, inlcuding DropEdge~\citep{rong2020dropedge}, and Graphon-based edge perturbation~\citep{hu2021training}. DropEdge removes graph edges independently with a specified probability, aiming to prevent over-smoothing and over-fitting issues in GNNs. Graphon-based edge perturbation~\citep{rong2020dropedge} improves the Dropedge by dropping edge based on an estimated probability. One limitation of such methods is that the edge permutation is based on one individual graph, so the graphs will not mix up.
DropEdge and Graphon-based edge perturbation~\citep{hu2021training} are special cases of $\mathcal{G}$-Mixup while setting different hyperparameter $\lambda$. \emph{i) $\mathcal{G}$-Mixup will degenerate into Graphon-based edge perturbation}, if $\lambda=0$ in \cref{equ:gmixup:mixw}, where the mathematical expression is ${W}_{\mathcal{I}} = {W}_{\mathcal{H}}, \{{I}_1, {I}_2, \cdots ,{I}_m\}  \stackrel{\text{i.i.d}}{\sim} \mathbb{G}(k, {W}_{\mathcal{I}}), \mathbf{y}_{\mathcal{I}} = \mathbf{y}_{\mathcal{H}}$.
\emph{ii) $\mathcal{G}$-Mixup will degenerate into DropEdge}, if $\lambda=0$ and using the element-wise product of graphons ${W}$ and adjacency matrix $\mathbf{A}$ in \cref{equ:gmixup:mixw} as edge probability. The expression is ${W}_{\mathcal{I}} = \mathbf{A} \odot {W}_{\mathcal{H}}, \{{I}_1, {I}_2, \cdots ,{I}_m\}  \stackrel{\text{i.i.d}}{\sim} \mathbb{G}(k, {W}_{\mathcal{I}}), \mathbf{y}_{\mathcal{I}} = \mathbf{y}_{ \mathcal{H} }$, where $\odot$ is element-wise multiplication.

\textbf{Relation to Manifold Mixup}. As a model-agnostic augmentation method, $\mathcal{G}$-Mixup has broader applications, e.g., creating graphs for graph contrastive learning, than Manifold Mixup. Manifold Mixup~\cite{wang2021mixup} is proposed to mix up graphs in the embedding space, which interpolates hidden representations of graphs. Interpolating hidden representation could limit its applications because: 1) algorithms must have hidden representation of graphs, and 2) models must be modified to adapt it. In contrast, $\mathcal{G}$-Mixup generates synthetic graphs without modifying models.

\section{Experiments}\label{sec:exp}
We evaluate the performance of $\mathcal{G}$-Mixup in this section. First, we visualize graphons and graph generation results in \cref{sec:exp:gon,sec:exp:vis} to investigate what $\mathcal{G}$-Mixup actually do on real-world datasets. Then, we evaluate the effectiveness of $\mathcal{G}$-Mixup in graph classification with various datasets and GNN backbones in \cref{sec:exp:perf}, as well as how it improves the robustness of GNNs against label corruption and adversarial attacks in \cref{sec:exp:robu}. The experiment setting and more experiments are in \cref{sec:appe:exp_detail,sec:appe:exp}. The observations are highlighted with \Circled{\footnotesize \#} \textbf{boldface}.

\subsection{Do different classes of real-world graphs have different graphons?}\label{sec:exp:gon}
We visualize the estimated graphons in \cref{fig:gon}.
It shows that, \Circled{\footnotesize 1}~\textbf{the graphons of different class of graphs in one dataset are distinctly different}. The graphons of IMDB-BINAERY in \cref{fig:gon} shows that the graphon of class $1$ has larger dense area, which indicates that the graphs in this class have a more large communities than the graphs of class $0$. The graphons of REDDIT-BINARY in \cref{fig:gon} shows that graphs of class $0$ have one high-degree nodes while the graphs of class $1$ have two. This observation validates that real-world graphs of different classes have distinctly different graphons, which lays a solid foundation for generating the mixture of graphs by mixing up graphons.

\begin{figure*}[tb]
 \centering
 \includegraphics[width=1\textwidth]{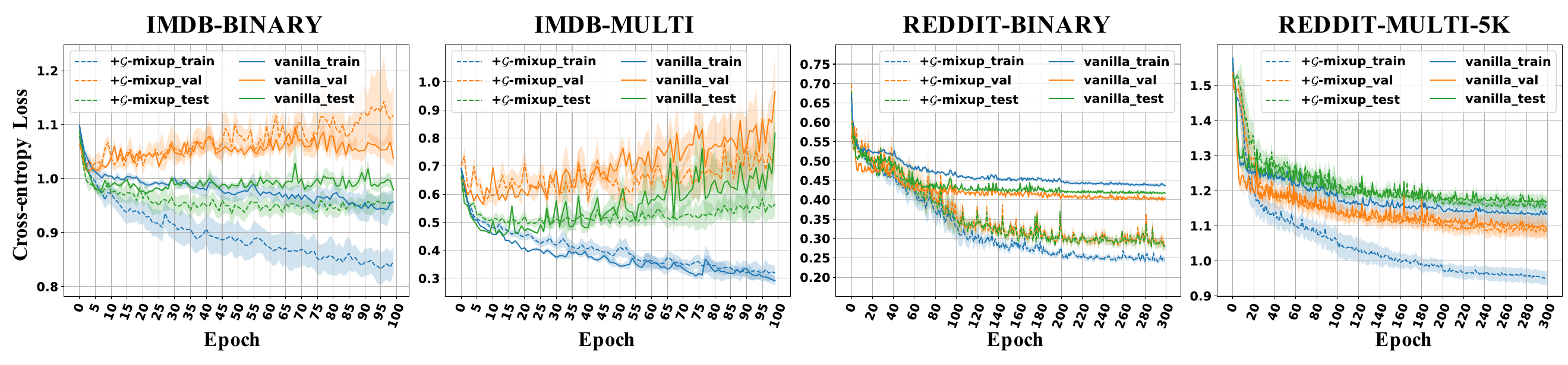}
  \vspace{-20pt} 
  \caption{The training/validation/test curves on IMDB-BINARY, IMDB-MULTI, REDDIT-BINARY and REDDIT-MULTI-5K with GCN as backbone. The curves are depicted on ten runs.}\label{fig:curve}
 \vspace{-25pt} 
\end{figure*}

\subsection{What is \texorpdfstring{$\mathcal{G}$}~-Mixup doing? A case study}\label{sec:exp:vis}
To investigate the outcome of $\mathcal{G}$-Mixup in real-world scenarios, we visualize the generated synthetic graphs in REDDIT-BINARY dataset in \cref{fig:sampling}. We observed that \Circled{\footnotesize 2}~\textbf{The synthetic graphs are indeed the mixture of the original graphs.} Original graphs and the generated synthetic graphs are visualized in \cref{fig:sampling}(a)(b) and \cref{fig:sampling}(c)(d)(e), respectively. \cref{fig:sampling} demonstrates that mixed graphon $0.5*{W}_{\mathcal{G}}+0.5*{W}_{\mathcal{H}}$ is able to generate graphs with a high-degree node and a dense subgraph, which can be regarded as the mixture of graphs with one high-degree node and two high-degree nodes. It validates that $\mathcal{G}$-Mixup prefer to preserve the discriminative motifs from the original graphs.

\begin{table}[!tp]
\centering
\fontsize{7}{8}\selectfont  
\setlength\tabcolsep{2pt}
\caption{Performance comparisons of $\mathcal{G}$-Mixup with different GNNs on different datasets. The metric is the classification accuracy. Experimental settings are in \cref{sec:appe:exp_detail}.} \label{tab:performance}
\begin{tabular}{ll|cccccccccc}
    \toprule
    \multicolumn{2}{c|}{Dataset} &
    \multicolumn{1}{c}{IMDB-B} &
    \multicolumn{1}{r}{IMDB-M} & 
    \multicolumn{1}{r}{REDD-B} & 
    \multicolumn{1}{r}{REDD-M5} &
    \multicolumn{1}{r}{REDD-M12} \\
 \midrule
 \multicolumn{2}{c|}{ \#graphs  }       &1000  &1500    &2000  &4999 &11929       \\
 \multicolumn{2}{c|}{ \#classes  }      &2   &3       &2      &5    &11     \\
 \multicolumn{2}{c|}{ \#avg.nodes  }    &19.77   &13.00       &429.63      &508.52    &391.41     \\
 \multicolumn{2}{c|}{ \#avg.edges  }    &96.53   &65.94       &497.75      &594.87    &456.89     \\

 \midrule[0.8pt]
 \parbox[t]{2mm}{\multirow{3}{*}{\rotatebox[origin=c]{90}{GCN}}}  &vanilla                  &\ms{72.18}{1.55}    &\ms{48.79}{2.72}  &\ms{78.82}{1.33}   &\ms{45.07}{1.70}    &\ms{46.90}{0.73}     \\
                            &w/ Dropedge              &\ms{72.50}{0.31}    &\ms{49.08}{1.89}  &\ms{81.25}{8.15}   &\ms{51.35}{1.54}    &\ms{47.08}{0.55}    \\
                            &w/ DropNode          &\ms{72.00}{4.09}  &\ms{48.58}{2.85}  &\ms{79.25}{0.35}  &\ms{49.35}{1.80}  &\ms{47.93}{0.64}\\
                            &w/ Subgraph              &\ms{68.50}{4.76}  &\ms{49.58}{2.61}  &\ms{74.33}{2.88}  &\ms{48.70}{1.63}  &\ms{47.49}{0.93} \\
                            &w/ M-Mixup          &\ms{72.83}{1.75}  &\ms{49.50}{1.97}  &\ms{75.75}{4.53}  &\ms{49.82}{0.85}  &\ms{46.92}{1.05} \\
                            &w/ $\mathcal{G}$-Mixup   &\bms{72.87}{3.85}   &\bms{51.30}{2.14} &\bms{89.81}{0.74}  &\bms{51.51}{1.70}   &\bms{48.06}{0.53}   \\
 \midrule

\parbox[t]{2mm}{\multirow{3}{*}{\rotatebox[origin=c]{90}{GIN}}}       &vanilla                   &\ms{71.55}{3.53}   &\ms{48.83}{2.75}   &\ms{92.59}{0.86}   &\ms{55.19}{1.02}   &\ms{50.23}{0.83}      \\
                            &w/ Dropedge               &\bms{72.20}{1.82}   &\ms{48.83}{3.02}   &\ms{92.00}{1.13}   &\ms{55.10}{0.44}   &\ms{49.77}{0.76}\\
                            &w/ DropNode           &\ms{72.16}{0.28} &\ms{48.33}{0.98} &\ms{90.25}{0.98} &\ms{53.26}{4.99} &\ms{49.95}{1.70}\\
                            &w/ Subgraph               &\ms{68.50}{0.86} &\ms{47.25}{3.78} &\ms{90.33}{0.87} &\ms{54.60}{3.15} &\ms{49.67}{0.90}\\
                            &w/ M-Mixup           &\ms{70.83}{1.04} &\ms{49.88}{1.34} &\ms{90.75}{1.78} &\ms{54.95}{0.86} &\ms{49.81}{0.80}\\
                            &w/ $\mathcal{G}$-Mixup    &\ms{71.94}{3.00}  &\bms{50.46}{1.49}  &\bms{92.90}{0.87}  &\bms{55.49}{0.53} &\bms{50.50}{0.41}    \\
\bottomrule
\end{tabular}
\vspace{-20pt}
\end{table}

\subsection{Can \texorpdfstring{$\mathcal{G}$}~-Mixup improve the performance and generalization of GNNs?}\label{sec:exp:perf}

To validate the effectiveness of $\mathcal{G}$-Mixup, we compare the performance of GNNs with various backbones on different datasets, and summarize results in \cref{tab:performance,tab:mor_gnns} as well as the training curves in \cref{fig:curve}. We make the following observations:
\Circled{\footnotesize 3}~\textbf{$\mathcal{G}$-Mixup can improve the performance of graph neural networks on various datasets.} From \cref{tab:performance}, $\mathcal{G}$-Mixup gain $12$ best performances among $15$ reported accuracies, which substantially improve the performance of GNNs. Overall, $\mathcal{G}$-Mixup performs 2.84\% better than the vanilla model. Note that $\mathcal{G}$-Mixup and baseline models adopt the same architecture of GNNs (e.g., layers, activation functions) and the same training hyperparameters (e.g., optimizer, learning rate). From \cref{tab:mor_gnns}, $\mathcal{G}$-Mixup gains $7$ best performances among $8$ cases, which substantially improve the performance of DiffPool and MincutPool.
Meanwhile, \Circled{\footnotesize 4}~\textbf{$\mathcal{G}$-Mixup can improve the generalization of graph neural networks.} From the loss curve on test data (green line) in \cref{fig:curve}, the loss of test data of $\mathcal{G}$-Mixup (dashed green lines) are consistently lower than the vanilla model (solid green lines). Considering both the better performance and the better test loss curves, $\mathcal{G}$-Mixup is able to substantially improve the generalization of GNNs. Also, \Circled{\footnotesize 5}~\textbf{$\mathcal{G}$-Mixup could stabilize the model training.} As shown in \cref{tab:performance}, $\mathcal{G}$-Mixup achieves $11$ lower standard deviation among total $15$ reported numbers than the vanilla model. Additionally, the train/validation/test curves of $\mathcal{G}$-Mixup (dashed line) in \cref{fig:curve} are more stable than vanilla model (solid line), indicating that $\mathcal{G}$-Mixup stabilize the training of graph neural networks. Experiments on OGB~\citep{hu2020open} and more pooling method (GMT) are in \cref{sec:appe:gmt,sec:appe:ex_ogb}.

\subsection{Can \texorpdfstring{ $\mathcal{G}$ }~-Mixup improve the robustness of GNNs?}\label{sec:exp:robu}
We investigate the two kinds of robustness of $\mathcal{G}$-Mixup, including \emph{Label Corruption Robustness} and \emph{Topology Corruption Robustness}, and report the results in \cref{tab:expe:labelnoise} and \cref{tab:expe:attack}, respectively. More experimental settings are presented in \cref{sec:appe:roub}. \Circled{\footnotesize 6}~\textbf{$\mathcal{G}$-Mixup improves the robustness of graph neural networks.} \cref{tab:expe:labelnoise} shows $\mathcal{G}$-Mixup gains better performance in general, indicating it is more robust to noisy labels than the vanilla baseline. \cref{tab:expe:attack} shows that $\mathcal{G}$-Mixup is more robust when graph topology is corrupted since the accuracy is consistently better than baselines. This can be an advantage of $\mathcal{G}$-Mixup when graph label or topology are noisy.

\begin{table}[!tp]
\centering
\fontsize{7}{8}\selectfont
\setlength\tabcolsep{5pt}
\caption{Performance comparisons of $\mathcal{G}$-Mixup with different Pooling methods. The metric is classification accuracy.}\label{tab:mor_gnns}
\begin{tabular}{llrrrrrrrrrr}
    \toprule
    &
    \multicolumn{1}{l}{Method} &
    \multicolumn{1}{c}{IMDB-B} &
    \multicolumn{1}{c}{IMDB-M} & 
    \multicolumn{1}{c}{REDD-B} & 
    \multicolumn{1}{c}{REDD-M5k} \\

 \midrule
\parbox[t]{2mm}{\multirow{3}{*}{\rotatebox[origin=c]{90}{TopKPool}}}  &vanilla                   &\ms{72.37}{5.01}    &\ms{50.57}{1.62}    &\ms{90.30}{1.47}     &\ms{45.07}{1.70}  \\
                            &w/ Dropedge               &\ms{71.75}{2.18}    &\ms{48.75}{2.94}    &\ms{88.96}{1.90}    &\bms{47.43}{1.82}  \\
                            &w/ DropNode           &\ms{69.16}{1.04} &\ms{48.50}{2.50} &\ms{81.33}{4.48} &\ms{46.15}{2.28} \\
                            &w/ Subgraph               &\ms{67.83}{4.01} &\ms{50.83}{2.38} &\ms{86.08}{2.12} &\ms{45.75}{2.47} \\
                            &w/ M-Mixup           &\ms{71.83}{3.03} &\ms{51.22}{1.17} &\ms{87.58}{3.16} &\ms{45.60}{2.35} \\
                            &w/ $\mathcal{G}$-Mixup    &\bms{72.80}{3.33}     &\bms{51.30}{2.14}  &\bms{90.40}{0.89}    &\ms{46.48}{1.70}  \\
 \midrule
\parbox[t]{2mm}{\multirow{3}{*}{\rotatebox[origin=c]{90}{DiffPool}}}  &vanilla                  &\ms{71.68}{3.40}  &\ms{47.75}{2.34}  &\ms{78.40}{4.38}  &\ms{31.61}{5.95}\\
                            &w/ Dropedge              &\ms{69.16}{2.51}  &\ms{49.44}{2.50}  &\ms{76.00}{5.50}  &\ms{34.46}{6.80} \\
                            &w/ DropNode          &\ms{70.25}{3.01}  &\ms{46.83}{1.34}  &\ms{76.68}{2.57}  &\ms{33.10}{5.53}\\
                            &w/ Subgraph              &\ms{69.50}{2.16}  &\ms{46.00}{4.43}  &\ms{76.06}{2.81}  &\ms{31.65}{4.43} \\
                            &w/ M-Mixup          &\ms{66.50}{4.04}  &\ms{45.16}{4.63}  &\ms{78.37}{2.29}  &\ms{34.46}{6.80} \\
                            &w/ $\mathcal{G}$-Mixup   &\bms{73.25}{3.89}  &\bms{50.70}{2.79}  &\bms{78.87}{2.27}  &\bms{38.42}{6.51} \\
 \midrule
\parbox[t]{2mm}{\multirow{3}{*}{\rotatebox[origin=c]{90}{MincutPool}}} &vanilla                    &\ms{73.25}{3.27}  &\ms{49.04}{3.57}  &\ms{84.95}{3.25}  &\ms{49.32}{2.67}  \\
                            &w/ Dropedge                &\ms{69.16}{2.51}  &\ms{49.66}{1.73}  &\ms{81.37}{1.59}  &\ms{47.20}{1.10}  \\
                            &w/ DropNode            &\ms{73.50}{3.89}  &\ms{49.91}{2.83}  &\ms{85.68}{2.04}  &\ms{46.82}{4.60}  \\
                            &w/ Subgraph                &\ms{70.25}{1.84}  &\ms{48.18}{1.10}  &\ms{84.91}{2.50}  &\ms{49.22}{2.49}   \\
                            &w/ M-Mixup            &\ms{70.62}{2.09}  &\ms{49.96}{1.86}  &\ms{85.12}{2.29}  &\ms{47.20}{1.10} \\
                            &w/ $\mathcal{G}$-Mixup     &\bms{73.93}{2.84}  &\bms{50.29}{2.30}  &\bms{85.87}{1.37}  &\bms{50.12}{2.47}  \\
\bottomrule
\end{tabular}
\vspace{-20pt}
\end{table}

\begin{table}[!tp]
\centering
\scriptsize
\setlength\tabcolsep{3pt}
\fontsize{7.2}{8}\selectfont  
\caption{Robustness to label corruption with different ratios.}\label{tab:expe:labelnoise}

\begin{tabular}{llrrrrrrrrrrr}
    \toprule
    \multicolumn{1}{l}{Models} &
    \multicolumn{1}{l}{Methods} &
    \multicolumn{1}{c}{10\%} &
    \multicolumn{1}{c}{20\%} &
    \multicolumn{1}{c}{30\%} &
    \multicolumn{1}{c}{40\%}  \\

 \midrule
 \multirow{1}{*}{IMDB-B}    &vanilla                   &\ms{72.30}{3.67}      &\ms{69.43}{4.80}     &\ms{63.65}{8.87}   &\bms{55.21}{8.75}   \\
                            &w/ Dropedge               &\ms{72.00}{2.44}     &\ms{69.52}{3.25}    &\ms{64.12}{3.44}  &\ms{48.50}{0.00} \\
                            &w/ M-Mixup           &\ms{71.87}{3.56}     &\ms{69.03}{4.85}    &\ms{65.62}{9.89}  &\ms{48.50}{0.00} \\
                            &w/ $\mathcal{G}$-Mixup    &\bms{72.56}{3.08}     &\bms{69.87}{5.41}    &\bms{65.50}{8.90}  &\ms{52.56}{6.97}  \\

 \midrule
 \multirow{1}{*}{REDD-B}  &vanilla                    &\bms{73.90}{1.43}  &\ms{75.68}{2.75}   &\ms{68.12}{0.81}   &\ms{46.50}{0.00}   \\
                            &w/ Dropedge                &\ms{73.75}{1.28}  &\ms{72.06}{1.42}  &\ms{46.50}{0.00}  &\ms{46.50}{0.00} \\
                            &w/ M-Mixup            &\ms{71.96}{1.97}  &\ms{76.00}{2.24}  &\ms{54.43}{1.09}  &\ms{46.50}{0.00} \\
                            &w/ $\mathcal{G}$-Mixup     &\ms{71.94}{3.00}   &\bms{76.34}{1.49}  &\bms{74.21}{1.85}  &\bms{53.50}{0.00}  \\
\bottomrule
\end{tabular}

\vspace{-20pt}
\end{table}

\begin{table}[!tp]
\centering
\scriptsize
\setlength\tabcolsep{3pt}
\fontsize{7}{8}\selectfont  
\caption{Robustness to topology corruption with different ratios.}\label{tab:expe:attack}

\begin{tabular}{llrrrrrrrrr}
    \toprule
    \multicolumn{1}{l}{Models} &
    \multicolumn{1}{l}{Methods} &
    \multicolumn{1}{c}{10\%} &
    \multicolumn{1}{c}{20\%} &
    \multicolumn{1}{c}{30\%} &
    \multicolumn{1}{c}{40\%} \\

 \midrule
 \multirow{1}{*}{Removing}             &vanilla                     &\ms{77.96}{3.71}      &\ms{67.59}{5.73}     &\ms{64.96}{8.87}   &\ms{65.71}{8.31}    \\
 \multirow{1}{*}{edges}                &w/ Dropedge                 &\ms{74.40}{2.26}      &\ms{65.12}{3.51}      &\ms{65.93}{2.32}  &\ms{57.87}{4.14} \\
                                       &w/ M-Mixup             &\ms{75.62}{1.59}      &\ms{65.81}{3.84}      &\ms{59.81}{9.45}  &\ms{57.31}{3.15} \\
                                       &w/ $\mathcal{G}$-Mixup      &\bms{81.46}{3.08}     &\bms{71.12}{7.47}    &\bms{67.46}{8.90}  &\bms{66.25}{7.78}  \\

 \midrule
 \multirow{1}{*}{Adding}    &vanilla                &\ms{76.12}{5.73}  &\ms{74.37}{6.48}  &\ms{72.31}{2.69}   &\ms{72.00}{2.92}  \\
 \multirow{1}{*}{edges}     &w/ Dropedge                &\ms{70.53}{1.47} &\ms{70.18}{1.29} &\ms{71.18}{1.53}  &\ms{70.90}{1.53} \\
                            &w/ M-Mixup            &\ms{73.41}{2.40} &\ms{71.87}{1.28} &\ms{71.50}{2.03}  &\ms{71.21}{2.00} \\
                            &w/ $\mathcal{G}$-Mixup     &\bms{84.31}{3.21} &\bms{82.21}{4.31} &\bms{77.00}{2.25}  &\bms{75.56}{3.05}\\
\bottomrule
\end{tabular}
\vspace{-10pt}
\end{table}

\subsection{Further Analysis}\label{sec:appe:futh}

\subsubsection{The nodes number of generated graphs}
\begin{figure}[!tp]
 \centering
 \includegraphics[width=0.48\textwidth]{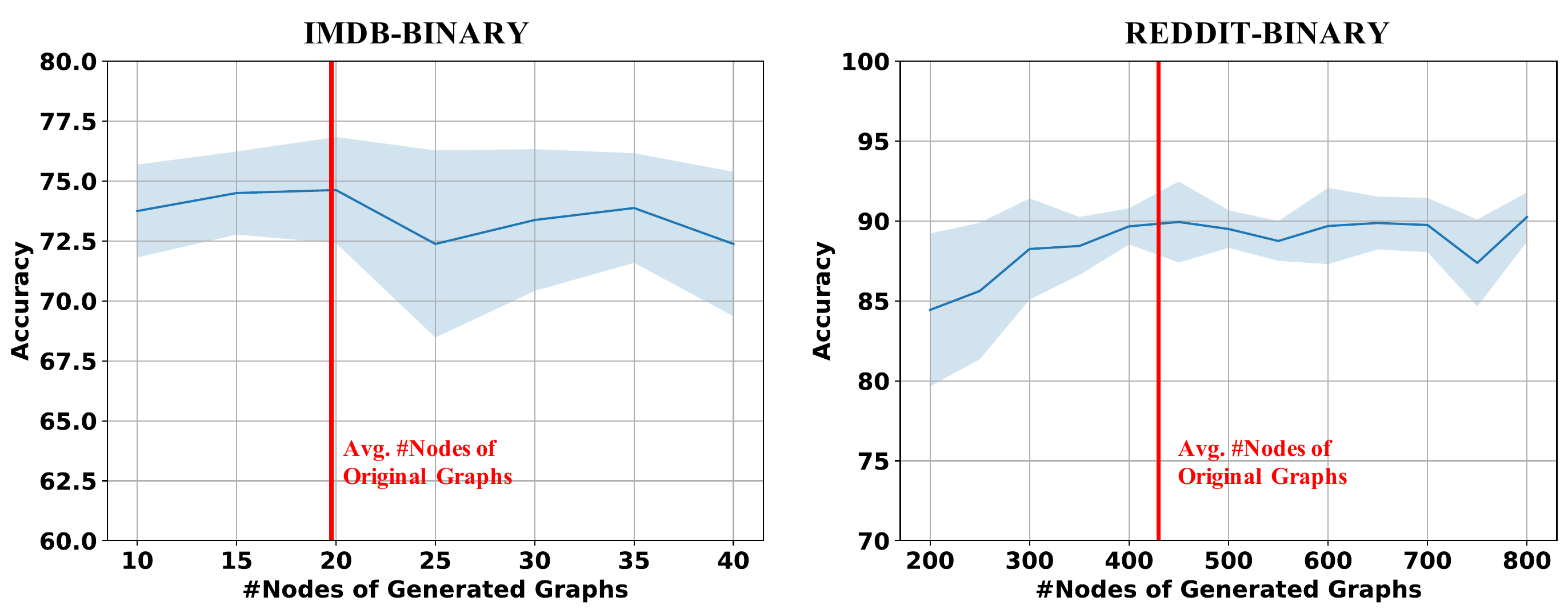}
  \vspace{-25pt} 
  \caption{The impact of the node numbers of generated synthetic graphs. The \textcolor{red}{red} vertical line indicates the average number of all the original training graphs. The \textcolor{blue}{blue} line represents that classification accuracy with different number of nodes of generated graphs.}\label{fig:node_num}
  \vspace{-10pt} 
\end{figure}

We investigate the impact of the nodes number in generated synthetic graphs by $\mathcal{G}$-Mixup and present the results in \cref{fig:node_num}. Specifically, $\mathcal{G}$-Mixup generates synthetic graphs with different numbers (hyperparameters $K$) of nodes and use them to train graph neural networks. We observed form~\cref{fig:node_num} that \Circled{\footnotesize 7}~\textbf{using the average node number of all the original graphs is a better choice for hyperparameter $K$ in $\mathcal{G}$-Mixup}, which is in line with the intuition.

%  \vspace{-20pt} 

\subsubsection{Impact on Deeper Models}
\begin{figure}[!tp]
 \centering
 \includegraphics[width=0.48\textwidth]{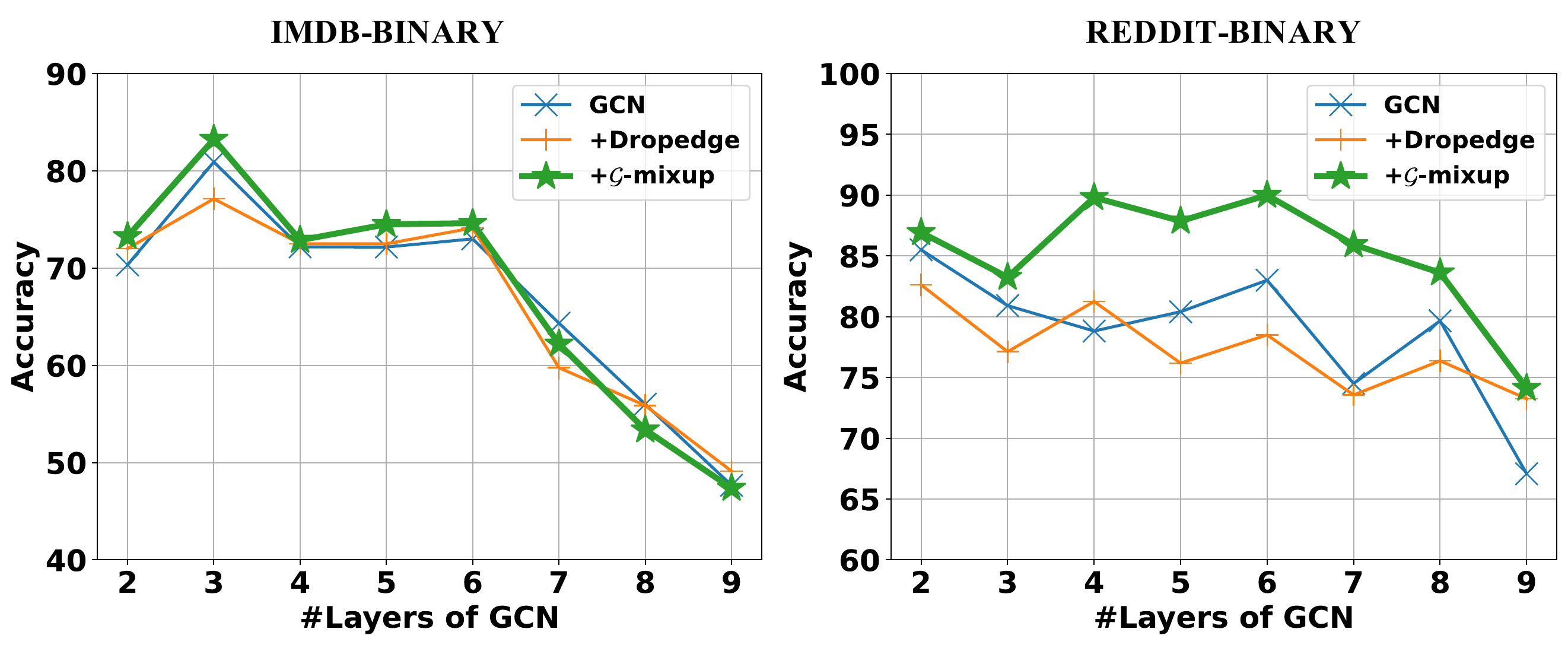}
  \vspace{-25pt} 
  \caption{The performance of $\mathcal{G}$-Mixup using GCNs with different layers on IMDB-BINARY and REDDIT-BINARY.}\label{fig:layers}
   \vspace{-10pt} 
\end{figure}

We investigate the performance of $\mathcal{G}$-Mixup when GCN goes deeper. We experiment with different numbers ($2-9$) of layers and report the results in \cref{fig:layers}. \Circled{\footnotesize 8}~\textbf{$\mathcal{G}$-Mixup improves the performance of graph neural networks with varying layers}. In \cref{fig:layers}, the left figure shows $\mathcal{G}$~-Mixup gains better performance while the depth of GCNs is $2-6$. The performance with deeper GCNs ($7-9$) are comparable to baselines, however, the accuracy is much lower than shallow ones. The right figure shows $\mathcal{G}$-Mixup gains better performance by a significant margin while the depth of GCNs is $2-9$. This validates the effectiveness of $\mathcal{G}$-Mixup when graph neural network goes deeper.
% \vspace{-20pt} 

\section{Related Works}\label{sec:rela}
\textbf{Graph Data Augmentation.}~
Graph neural networks (GNNs) achieve the state-of-the-art performance on graph classification tasks~\citep{kipf2016semi, velivckovic2017graph,hamilton2017inductive,xu2018powerful,zhang2018end}. 
In parallel, graph data augmentation methods improve the performance of GNNs. There are three categories of graph data augmentation, including node perturbation~\citep{you2020graph, huang2018adaptive}, edge perturbation~\citep{rong2020dropedge, you2020graph}, and subgraph sampling~\citep{you2020graph,wang2020graphcrop}. However, the major limitation of the existing graph augmentation methods is that they are based on one single graph while $\mathcal{G}$-Mixup leverages multiple input graphs. Besides, there are a line of works focusing on graph data augmentation methods for node classification~\citep{zhao2021data,wang2020nodeaug,tang2021data,park2021metropolis,verma2019graphmix}. The more discussion are in \cref{sec:appe:node}.

\textbf{Graphon Estimation.}~~~Graphons and convergent graph sequences have been broadly studied in mathematics \citep{lovasz2012large,lovasz2006limits,borgs2008convergent} and have been applied to network science~\citep{avella2018centrality,vizuete2021laplacian} and graph neural networks~\citep{ruiz2020graphon, ruiz2020graph}. 
There are tow lines of works to estimate step functions, one is based on stochastic block models, such as stochastic block approximation (SBA)~\citep{airoldi2013stochastic}, ``largest gap'' (LG)~\citep{channarond2012classification} and sorting-and-smoothing (SAS)~\citep{chan2014consistent}; another one is based on low-rank matrix decomposition, such as matrix completion (MC)~\citep{keshavan2010matrix}, universal singular value thresholding (USVT)~\citep{chatterjee2015matrix}. More discussion about graphon estimation are in \cref{sec:appe:gonm}.

\section{Conclusion}\label{sec:conlcu}
This work develops a novel graph augmentation method called $\mathcal{G}$-Mixup. Unlike image data, graph data is irregular, unaligned and in non-Euclidean space, making it hard to be mixed up. However, the graphs within one class have the same generator (i.e., graphon), which is regular, well-aligned and in Euclidean space. Thus we turn to mix up the graphons of different classes to generate synthetic graphs. $\mathcal{G}$-Mixup is mix up and interpolate the topology of different classes of graphs. Comprehensive experiments show that GNNs trained with $\mathcal{G}$-Mixup achieve better performance and generalization, and improve the model robustness to noisy labels and corrupted topology.

% \section*{Reproducibility}
% Authors are kindly asked to make their submissions as accessible as possible for everyone including people with disabilities and sensory or neurological differences.
% Tips of how to achieve this and what to pay attention to will be provided on the conference website \url{http://icml.cc/}.


\bibliography{paper}
\bibliographystyle{icml2022}

%%%%%%%%%%%%%%%%%%%%%%%%%%%%%%%%%%%%%%%%%%%%%%%%%%%%%%%%%%%%%%%%%%%%%%%%%%%%%%%
%%%%%%%%%%%%%%%%%%%%%%%%%%%%%%%%%%%%%%%%%%%%%%%%%%%%%%%%%%%%%%%%%%%%%%%%%%%%%%%
% APPENDIX
%%%%%%%%%%%%%%%%%%%%%%%%%%%%%%%%%%%%%%%%%%%%%%%%%%%%%%%%%%%%%%%%%%%%%%%%%%%%%%%
%%%%%%%%%%%%%%%%%%%%%%%%%%%%%%%%%%%%%%%%%%%%%%%%%%%%%%%%%%%%%%%%%%%%%%%%%%%%%%%
\newpage
\appendix
\onecolumn

\section{Proof of Theorem}\label{sec:appe:proo}

In the appendix, we first present the preliminaries in \cref{sec:appe:prel}. 
And then we present complete proof for \cref{theo:bound,theo:sample} in \cref{sec:appe:proof_c1,sec:appe:proof_p1}, respectively.

\subsection{Preliminaries}\label{sec:appe:prel}

Cut norm~\citep{lovasz2012large,zhao202graphcom} is used to measure structural similarity of two graphons. The definition of cut norm is as follow:
\begin{definition}
The cut norm of grapon $W$ is defined as
\begin{eqnarray}\label{eq:cut_n}
\begin{aligned}
\|W\|_{\square}=\sideset{}{_{\mathrm{S}, \mathrm{T} \subset [0,1] }}\sup\Bigl| \int_{\mathrm{S} \times \mathrm{T} } W(x, y)dxdy\Bigr|,
\end{aligned}
\end{eqnarray}
\end{definition}
where the supremum is taken over all measurable subsets $\mathrm{S}$ and $\mathrm{T}$.

The following lemma follows the derivation of counting lemma for graphons, are known in the paper~\citep{lovasz2012large}. It will be used to prove the \cref{theo:bound}.
\begin{lemma}\label{lemma:1}
Let $F$ be a simple graph and let $W, W^{\prime} \in \mathcal{W}$. Then
\begin{equation}
\begin{aligned}
     |t(F,W) - t(F,W^{\prime}) | \leq \mathrm{e}(F)|| W-W^{\prime} ||_{\square}
\end{aligned}
\end{equation}
\end{lemma}

\textit{Proof of \cref{lemma:1}:} The proof follows ~\citet{zhao202graphcom}. For an arbitrary simple graph $F$, by the triangle inequality we have
\begin{equation}
    \begin{aligned}
        &|t(F, W)-t(F, W')| \\
        &=\left|\int\left(\prod_{u_{i} v_{i} \in E} W\left(u_{i}, v_{i}\right)-\prod_{u_{i} v_{i} \in E} W'\left(u_{i}, v_{i}\right)\right) \prod_{v \in V} d v\right| \\
        &\leq \sum_{i=1}^{|E|}\left|\int\left(\prod_{j=1}^{i-1} W'\left(u_{j}, v_{j}\right)\left(W\left(u_{i}, v_{i}\right)-W'\left(u_{i}, v_{i}\right)\right) \prod_{k=i+1}^{|E|} W\left(u_{k}, v_{k}\right)\right) \prod_{v \in V} d v\right|
    \end{aligned}
\end{equation}

Here, each absolute value term in the sum is bounded by the cut norm $\| W - W' \|_{\square}$  if we fix all other irrelavant variables (everything except $u_{i}$ and $v_{i}$ for the $i$-th term), altogether implying that
\begin{equation}
    \mid t(F, W)-t(F, W')|\leq \mathrm{e}(F)|| W-W^{\prime} ||_{\square}
\end{equation}

$\hfill\blacksquare$

\begin{lemma}[Corollary 10.4 in~\cite{lovasz2006limits}]\label{lem:2}
Let ${W}$ be a graphon, $n\geq 1$, $0<\varepsilon<1$, and let $F$ be a simple graph, then the $W$-random graph $\mathbb{G}=\mathbb{G}(n, W)$ satisfies 
\begin{equation}
    \mathrm{P} \left( |t({F}, \mathbb{G} )-t( {F},{W} )|>\varepsilon \right) \leq 2\mathrm{exp}\left(-\frac{\varepsilon^2n}{8 \mathrm{v}(F)^2 } \right)
\end{equation}
\end{lemma}

\subsection{Proof of Theorem 1}\label{sec:appe:proof_p1}

We have the mixed graphon ${W}_{\mathcal{I}} = \lambda {W}_{\mathcal{G}} + (1-\lambda) {W}_{\mathcal{H}}$. Let $W = {W}_{\mathcal{I}}$, $W^{\prime} =  {W}_{\mathcal{G}} $, and $F = F_{\mathcal{G}}$ in \cref{lemma:1}, we have,
\begin{equation}\label{equ:1}
\begin{aligned}
    |t(F_{\mathcal{G}},{W}_{\mathcal{I}} ) - t(F_{\mathcal{G}},{W}_{\mathcal{G}} ) | &\leq \mathrm{e}(F_{\mathcal{G}})|| {W}_{\mathcal{I}} - {W}_{\mathcal{G}} ||_{\square}\\
    |t(F_{\mathcal{G}}, \lambda {W}_{\mathcal{G}} + (1-\lambda) {W}_{\mathcal{H}} ) - t(F,{W}_{\mathcal{G}} ) | &\leq \mathrm{e}(F_{\mathcal{G}})|| \lambda {W}_{\mathcal{G}} + (1-\lambda) {W}_{\mathcal{H}} - {W}_{\mathcal{G}} ||_{\square}\\
     &\leq  \mathrm{e}(F_{\mathcal{G}})|| (1-\lambda)({W}_{\mathcal{H}} - {W}_{\mathcal{G}}) ||_{\square}
\end{aligned}
\end{equation}

Recall that the cut norm $\|W\|_{\square}=\sup _{S, T \subseteq[0,1]}\left|\int_{S \times T} W\right| .$

obviously, suppose $\alpha \in \mathbb{R}$, we have
\begin{equation}\label{equ:2}
    \|\alpha W\|_{\square}=\sup _{S, T \subseteq[0,1]}\left|\int_{S \times T} \alpha W\right| = \sup _{S, T \subseteq[0,1]}\left|\alpha\int_{S \times T}  W\right|= \alpha\|W\|_{\square}
\end{equation}

Based on \cref{equ:1} and \cref{equ:2}, we have

\begin{equation}\label{equ:17}
\begin{aligned}
    |t(F_{\mathcal{G}}, \lambda {W}_{\mathcal{G}} + (1-\lambda) {W}_{\mathcal{H}} ) - t(F_{\mathcal{G}},{W}_{\mathcal{G}} ) |  &\leq  \mathrm{e}(F_{\mathcal{G}})|| (1-\lambda)({W}_{\mathcal{H}} - {W}_{\mathcal{G}}) ||_{\square}\\
    &\leq (1-\lambda) \mathrm{e}(F_{\mathcal{G}})|| {W}_{\mathcal{H}} - {W}_{\mathcal{G}} ||_{\square}
\end{aligned}
\end{equation}

Similarly, let $W = {W}_{\mathcal{I}}$, $W^{\prime} = {W}_{\mathcal{H}} $ and $F = F_{\mathcal{H}}$ in \cref{lemma:1}, We can also easily obtain
 
\begin{equation}\label{equ:18}
\begin{aligned}
    |t(F_{\mathcal{H}}, \lambda {W}_{\mathcal{G}} + (1-\lambda) {W}_{\mathcal{H}} ) - t(F_{\mathcal{H}},{W}_{\mathcal{H}} ) | &\leq \lambda \mathrm{e}(F_{\mathcal{H}})|| {W}_{\mathcal{H}} - {W}_{\mathcal{G}} ||_{\square} 
\end{aligned}
\end{equation}
\cref{equ:17} and~\cref{equ:18} produce the upper bound in \cref{equ:theo1}.$\hfill\blacksquare$

\subsection{Proof of Theorem 2} \label{sec:appe:proof_c1}
Let $F$ and $W$ be  the discriminative motif $F_{\mathcal{I}}$ and the mixed graphon ${W}_{\mathcal{I}}$ in \cref{lem:2}, we will have 
\begin{equation}
    \mathrm{P} \left( |t({F}_{\mathcal{I}}, \mathbb{G} )-t( {F}_{\mathcal{I}},{W}_{\mathcal{I}} )|>\varepsilon \right) \leq 2\mathrm{exp}\left(-\frac{\varepsilon^2n}{8 \mathrm{v}(F_{\mathcal{I}})^2 } \right)
    \vspace{-5pt}
\end{equation}

which produces the result in \cref{equ:theo2}.$\hfill\blacksquare$

\subsection{Graphons Estimation by Step Function}\label{sec:appe:step}
The proof follows \citet{xu2021learning}. A graphon can always be approximated by a step function in the cut norm~\citep{frieze1999quick}.

Let $\mathcal{P}=(\mathcal{P}_1,..,\mathcal{P}_K)$ be a partition of $\Omega$ into $K$ measurable sets. 
We define a step function $W_{\mathcal{P}}: \Omega^2\mapsto [0, 1]$ as
\begin{eqnarray}\label{eq:step}
\begin{aligned}
W_{\mathcal{P}}(x,y)=\sideset{}{_{k,k'=1}^{K}}\sum w_{kk'}1_{\mathcal{P}_k\times \mathcal{P}_{k'}}(x,y),
\end{aligned}
\end{eqnarray}
where each $w_{kk'}\in [0, 1]$ and the indicator function $1_{\mathcal{P}_k\times \mathcal{P}_{k'}}(x,y)$ is 1 if $(x, y)\in\mathcal{P}_{k}\times\mathcal{P}_{k'}$, otherwise it is 0. 
The weak regularity lemma~\cite{lovasz2012large} shown below guarantees that every graphon can be approximated well in the cut norm by step functions.
\begin{theorem}[Weak Regularity Lemma (Lemma 9.9 in ~\citep{lovasz2012large}) ]\label{thm:wrl}
For every graphon $W$ and $K\geq 1$, there always exists a step function $\mathbf{W}$ with $|\mathcal{P}|=K$ steps such that
\begin{eqnarray}\label{eeq:wrl}
\begin{aligned}
\|W - \mathbf{W}\|_{\square} \leq \frac{2}{\sqrt{\log K}}\|W\|_{L_2}.
\end{aligned}
\end{eqnarray}
\end{theorem}

\section{Graphons Estimation Methods}\label{sec:appe:gonm}
The adopted graphon estimated methods (e.g., LG, USVT, SBA) are well-studied methods. Typically they have rigorous mathematical proof to upper bound the graphon estimation error. For example, Theorem 2.10 in \citep{chatterjee2015matrix} shows the graphon estimation error of USVT is strictly upper bounded. And we also copy the results of graphon estimation methods on synthetic graphon from \citep{xu2021learning} in \cref{tab:gon_methods}. The results show the graphon estimation methods in our work can precisely estimate graphon. The details of them are listed as the following:

\begin{itemize}
    \item \textbf{SBA} \citep{airoldi2013stochastic} The Stochastic Block Approximation learns stochastic block models to approximate graphons. This method can consistently estimate the graphon with extremely small error and the estimation error vanishes provably as the node number of the graph goes infinity.
    \item \textbf{LG} \citep{channarond2012classification}  The “largest gap” algorithm improve the SBA method, which can be used for both large-scale and small graphs.
    \item \textbf{SAS} \citep{chan2014consistent}  The smoothing-and-sorting (SAS) is a improved variant of SBA, which first sorts the graphs based on the node degree, then smooths the sorted graph using total variation minimization.
    \item \textbf{MC and USVT}  \citep{keshavan2010matrix, chatterjee2015matrix} Matrix Completion and Universal Singular Value Thresholding are matrix decomposition based methods, which learn low-rank matrices to approximate graphons. 
\end{itemize}

\begin{table}
\centering
\setlength\tabcolsep{3pt}

\caption{The MSE error of graphon estimation methods on synthetic graphs\citep{xu2021learning}. The graphon estimation is based on $10$ graphs, the error is Mean Square Error, and the resolution of graphon is $1000\times 1000$.}
\label{tab:gon_methods}
\begin{tabular}{crrrrrr}
\toprule
\textbf{ $\mathbf{W}(x,y)$ }     &\textbf{ SBA } &\textbf{ LG } &\textbf{ MC } &\textbf{ USVT } &\textbf{ SAS }  \\
\midrule
$xy$                                    &65.6±6.5       &29.8±5.7      &11.3±0.8      &31.7±2.5        &125.0±1.3\\[5pt]
$e^{-(x^{0.7} + y^{0.7})}$              &58.7±7.8       &22.9±3.1      &71.7±0.5      &12.2±1.5        &77.7±0.8\\[5pt]
$\frac{x^2+y^2+\sqrt{x}+\sqrt{y} }{4}$   &63.4±7.6       &24.1±2.5      &73.2±0.7      &33.8±1.1        &99.3±1.2 \\[5pt]
$\frac{1}{2}(x+y)$                      &66.2±8.3       &24.0±2.5      &71.9±0.6      &40.2±0.8        &108.3±1.0\\[5pt]
$\frac{1}{1+exp(-10(x^2+y^2))}$         &5.0±9.5        &23.1±3.2      &64.6±0.5      &37.3±0.6        &73.3±0.7 \\[5pt]
\bottomrule
\end{tabular}
\end{table}

\section{Discussion about Manifold Intrusion in $\mathcal{G}$-Mixup}\label{sec:appe:intru}
In this appendix, we discuss that manifold intrusion in $\mathcal{G}$-Mixup and argue that $\mathcal{G}$-Mixup does not suffer from manifold intrusion issue. The manifold intrusion may be harmful for mixup method. Manifold intrusion in mixup is a form of under-fitting resulting from conflicts between the labels of the synthetic examples and the labels of original training data~\citep{guo2019mixup}. The manifold intrusion in graph learning represents that the generated graphs have identical topology but different labels. In our method, the adjacency matrix $\mathbf{A}\in \mathbb{R}^{K\times K}$ of generated graphs are generated from the matrix-from graphon $\mathbf{W}\in \mathbb{R}^{K\times K}$, thus we have $\mathbf{A}_{ij} \stackrel{\text{iid}}{\sim} \text{Bern}(\mathbf{W}_{ij}), \forall i,j \in [K]$. In the graph generation phase, $\mathcal{G}$-Mixup may cause manifold intrusion in two cases: 1) two generated two graphs are identical, 2) a generated graph is identical to an original graph. We hereby show that graph manifold intrusion issue will not happen with a very high probability in $\mathcal{G}$-Mixup as follows:
\begin{itemize}
\item \textbf{Two generated two graphs are identical.} The probability of generating two identical graphs from the same graphon $\mathbf{W}$ is $\Pi_{i=1}^{K}\Pi_{j=1}^{K}(\mathbf{W}_{ij}^2 + (1-\mathbf{W}_{ij})^2)$, which is extremely small since $0<\mathbf{W}_{ij}^2 + (1-\mathbf{W}_{ij})^2< 1$ and $K$ is large enough in the real-world graphs. The probability that two generated two graphs are identical are extremely small.
\item \textbf{A generated graph is identical to an original graph.} The probability of generating a new graph that is identical to an original graph (the adjacency matrix is $\tilde{\mathbf{A}}$) is $\Pi_{i=1}^{K}\Pi_{j=1}^{K}(\mathbf{W}_{ij}^{\tilde{\mathbf{A}}_{ij}}(1-\mathbf{W}_{ij})^{1-\tilde{\mathbf{A}}_{ij}})$, which is extremely small since $0<\mathbf{W}_{ij}^{\tilde{\mathbf{A}}_{ij}}(1-\mathbf{W}_{ij})^{1-\tilde{\mathbf{A}}_{ij}}< 1$ and $K$ is large enough in the real-world graphs. The probability that a generated graph is identical to an original graph are identical are extremely small too.
\end{itemize}

\section{More Discussion about Related Works}\label{sec:appe:node}
In this appendix, we discuss two categories of related works. The first one is graph data augmentation for node classification, and the second is model-dependent graph data augmentation for graph classification. Both of them are different to our proposed $\mathcal{G}$-Mixup. 

\textbf{Graph Data Augmentation for Node Classification.}~There is another line of works targeting graph data augmentation for node classification~\citep{zhao2021data, wang2020nodeaug, tang2021data,park2021metropolis,verma2019graphmix}. \citet{zhou2020data} leverage information inherent in the graph to predict edge probability to augment a new graph for node classification task. \citet{verma2019graphmix} proposed GraphMix to augment the vanilla GNN with a Fully-Connected Network (FCN) and the FCN loss is computed using Manifold Mixup. \citet{verma2019graphmix} proposed to generate augmented graphs from an explicit target distribution for semi-supervised learning, which has flexible control of the strength and diversity of augmentation. Many graph augmentation methods are proposed to solve node classificaiton task. However, the node classification task is a different task in graph learning from graph classification task. The node classification task usually has one input graph, thus the graph augmentation methods for node classification is limited to one graph while the graph augmentation for graph classification can manipulate multiple graphs. Thus graph data augmentation for node-level task is not applicable to our scenario.

\textbf{Model-Dependent Graph Data Augmentation for Graph Classification.}  There are some model-dependent graph augmentation methods \citep{suresh2021adversarial, you2022bringing, zhou2020data} for graph classification task. \citet{suresh2021adversarial} proposed to enable GNNs to avoid capturing redundant information during the training by optimizing adversarial graph augmentation strategies used in graph contrastive learning during the training phase. \citet{you2022bringing}  proposed to learn a continuous prior parameterized by a neural network from data during contrastive training, which is used to augment graph. The difference between our proposal and these methods is that $\mathcal{G}$-Mixup an general model-agnostic graph data augmentation methods for graph classification.

\section{Implementation Details}\label{sec:appe:imp_detail}
In this appendix, we present the pseudo code for $\mathcal{G}$-Mixup. We first present the pseudo code for graphon estimation in \cref{alg:ge}, which depicts how to generate the graphon and the node features. Since our proposed method is a model-agnostic method, which can be conducted before the model training. Then we present the pseudo code $\mathcal{G}$-Mixup. The graphon estimation is based on the one class of graphs, thus we can estimate on graphon using all the graphs in the same class or a random batch of graphs in the same class. On this basis, we have two version of concrete implementations: 1) estimating graphon on graphon using all the graphs in the same class (\cref{alg:gmixup}), 2) estimating graphon on graphon using a random batch of graphs in the same class (\cref{alg:gmixup_b}). The first implementation provide more accurate estimated graphons while the second encourages more diversity of the synthetic graphs. Note that all these two versions can be done as a pre-processing before model training.

\begin{algorithm}[!t]
    \caption{Graphon Estimation}\label{alg:ge}
    \begin{algorithmic}
        \STATE{{\bfseries Input:} graph set $\mathcal{G}$, graphon estimator $g$ }\hfill {\color{blue}\COMMENT{each graph $G$ has adjacency matrix $\mathbf{A}$ and node features matrix $\mathbf{X}$}}
        \STATE{\textbf{Init:} sorted  adjacency matrix set $\bar{\mathcal{A}}=\{\}$ }
        % \STATE{\textbf{Init:} initialize each $\omega_i$ independently.}
        \FOR{each graph $G$ in $\mathcal{G}$}
            \STATE{Calculate the degree of each nodes in $G$}
            \STATE{Calculate sorted adjacency matrix $\bar{\mathbf{A}}$ by sorting $\mathbf{A}$ based on the degree}
            \STATE{Calculate sorted node features matrix $\bar{\mathbf{X}}$ by sorting $\mathbf{X}$ based on the degree}
            \STATE{Add the sorted adjacency matrix $\bar{\mathbf{A}}$ to  $\bar{\mathcal{A}}$}
        \ENDFOR
        \STATE{Estimate step function $\mathbf{W}_{\mathcal{G}}$ with $\bar{\mathcal{A}}$ using $g$}. {\color{blue}\COMMENT{we use LG as $g$ in experiments}}
        \STATE{Obtain graphon node feature $\bar{\mathbf{X}}_{\mathcal{G}}$ by average pooling $\mathbf{X}$} {\color{blue}\COMMENT{we can use other pooling method (e.g., maxpooling)}}
        \STATE{ {\bfseries Return:} $\mathbf{W}_{\mathcal{G}}$, $\bar{\mathbf{X}}_{\mathcal{G}}$ } 
    \end{algorithmic}
\end{algorithm}

\begin{algorithm}[!t]
    \caption{$\mathcal{G}$-Mixup}\label{alg:gmixup}
    \begin{algorithmic}
        \STATE {\bfseries Input:} train graph set $\mathcal{S}$, graphon estimator $g$, mixup ratio $\lambda$, augmented ratio $\alpha$ {\color{blue}\COMMENT{$0< \lambda, \alpha < 1$} }
        \STATE{\textbf{Init:} Synthetic graph set $\mathcal{I}=\{\}$.}
        \STATE{Obtain two graph sets $\mathcal{G}$ and $\mathcal{H}$ with different labels $\mathbf{y}_{\mathcal{G}}$ and $\mathbf{y}_{\mathcal{H}}$}
        \STATE{Estimate ($\mathbf{W}_{\mathcal{G}}$, $\bar{\mathbf{X}}_{\mathcal{G}}$) and ($\mathbf{W}_{\mathcal{H}}$, $\bar{\mathbf{X}}_{\mathcal{H}}$) from $\mathcal{G}$ and $\mathcal{H}$ using \cref{alg:ge} }
        \STATE{Mix up step function $\mathbf{W}_{\mathcal{I}} = \lambda\mathbf{W}_{\mathcal{G}} + (1-\lambda)\mathbf{W}_{\mathcal{H}}$}
        \STATE{Mix up graphon node features $\bar{\mathbf{X}}_{\mathcal{I}} = \lambda\bar{\mathbf{X}}_{\mathcal{G}} + (1-\lambda)\bar{\mathbf{X}}_{\mathcal{H}}$}
        \STATE{Sample $\alpha\cdot|\mathcal{S}|$ synthetic graphs based on $\mathbf{W}_{\mathcal{I}}$ and $\bar{\mathbf{X}}_{\mathcal{I}}$ and add them to $\mathcal{I}$} {\color{blue}\COMMENT{$|\mathcal{I}|=\alpha\cdot|\mathcal{S}|$ after augmentation} }
        \STATE{ {\bfseries Return:} synthetic graph set $\mathcal{I}$ }
    \end{algorithmic}
\end{algorithm}

\begin{algorithm}[!t]
    \caption{$\mathcal{G}$-Mixup (batch)}\label{alg:gmixup_b}
    \begin{algorithmic}
        \STATE{\textbf{Input:} train graph set $\mathcal{S}$ with $B$ batches, graphon estimator $g$, mixup ratio $\lambda$, augmented ratio $\alpha$}{\color{blue}\COMMENT{$0< \lambda, \alpha < 1$} }
        \STATE{\textbf{Init:} Synthetic graph set $\mathcal{I}=\{\}$}
        \FOR{batch in $\mathcal{S}$}
        \STATE{Obtain two graph sets $\mathcal{G}$ and $\mathcal{H}$ with different labels $\mathbf{y}_{\mathcal{G}}$ and $\mathbf{y}_{\mathcal{H}}$}
        \STATE{Estimate ($\mathbf{W}_{\mathcal{G}}$, $\bar{\mathbf{X}}_{\mathcal{G}}$) and ($\mathbf{W}_{\mathcal{H}}$, $\bar{\mathbf{X}}_{\mathcal{H}}$) from $\mathcal{G}$ and $\mathcal{H}$ using \cref{alg:ge} }
        \STATE{Mix up step function $\mathbf{W}_{\mathcal{I}} = \lambda\mathbf{W}_{\mathcal{G}} + (1-\lambda)\mathbf{W}_{\mathcal{H}}$}
        \STATE{Mix up graphon node features $\bar{\mathbf{X}}_{\mathcal{I}} = \lambda\bar{\mathbf{X}}_{\mathcal{G}} + (1-\lambda)\bar{\mathbf{X}}_{\mathcal{H}}$}
        \STATE{Sample $\alpha\cdot|\mathcal{S}|/B$ synthetic graphs based on $\mathbf{W}_{\mathcal{I}}$ and $\bar{\mathbf{X}}_{\mathcal{I}}$ and add them to $\mathcal{I}$} {\color{blue}\COMMENT{$|\mathcal{I}|=\alpha\cdot|\mathcal{S}|$ after for loop ends} }
        % \COMMENT{}
        \ENDFOR

        \STATE{ {\bfseries Return:} synthetic graph set $\mathcal{I}$ }
    \end{algorithmic}
\end{algorithm}

\section{Experiments Details}\label{sec:appe:exp_detail}

\subsection{Experimental Setting}\label{sec:appe:setting}
To ensure a fair comparison, we use the same hyperparater for modeling training and the same architecture for vanilla model and other baselines. For model training, we use the Adam optimizer\citep{kingma2015adam}. The initial learning rate is $0.01$ and will drop the learning rate by half every 100 epochs. The batch size is set to $128$. We split the dataset into train/val/test data by $7:1:2$. Note that best test epoch is selected on a validation set, and we report the test accuracy on ten runs. For hyperparemeter in $\mathcal{G}$-Mixup, we generate 20\% more graph for training graph. The graphons are estimated based on the training graphs. We use different $\lambda \in [0.1,0.2]$ to mix up the graphon and generate synthetic with different strength of mixing up. 

\subsection{Architectures of Graph Neural Networks}\label{sec:appe:gnns}
We adopted two categories of graph neural networks as our baselines, The first category is Graph Convolutional Network (GCN) and Graph Isomorphism Network (GIN). The second category is graph polling methods, including TopK Pooling (TopKPool), Differentiable Pooling (DiffPool), MinCut Pooling (MincutPool) and Graph Multiset Pooling (GMT). The details of the GNNs are listed as follows:
\begin{itemize}
    \item \textbf{GCN}\footnote{\url{https://github.com/pyg-team/pytorch_geometric/blob/1.7.2/examples/gcn2_ppi.py}}~\citep{kipf2016semi}. Four GNN layers and global mean pooling are applied. All the hidden units is set to 64. The activation is ReLU~\citep{nair2010rectified}.
    \item \textbf{GIN}\footnote{\url{https://github.com/pyg-team/pytorch_geometric/blob/1.7.2/examples/mutag_gin.py}}~\citep{xu2018powerful}. We apply five GNN layers and all MLPs have two layers. Batch normalization~\citep{ioffe2015batch} is applied on every hidden layer. All hidden units are set to 64. The activation is ReLU~\citep{nair2010rectified}.
    \item \textbf{TopKPool}\footnote{\url{https://github.com/pyg-team/pytorch_geometric/blob/1.7.2/examples/proteins_topk_pool.py}}~\citep{gao2019graph}. Three GNN layers and three TopK pooling are applied. A there-layer percetron are adopted to predict the labels. All the hidden units is set to 64. The activation is ReLU~\citep{nair2010rectified}.
    \item \textbf{DiffPool}\footnote{\url{https://github.com/pyg-team/pytorch\_geometric/blob/1.7.2/examples/proteins\_diff\_pool.py}}~\citep{ying2018hierarchical} is a differentiable graph pooling methods that can be adapted to various GNN architectures, which maps nodes to clusters based on their learned embeddings.
    \item \textbf{MincutPool}\footnote{\url{https://github.com/pyg-team/pytorch\_geometric/blob/1.7.2/examples/proteins\_mincut\_pool.py}}~\citep{bianchi2020spectral} is a differentiable pooling baselines. It learns a clustering function that can be quickly evaluated on out-of-sample graphs. 
    \item \textbf{GMT}\footnote{\url{https://github.com/JinheonBaek/GMT}}~\citep{baek2020accurate} is a multi-head attention based global pooling layer to generate graph representation, which captures the interaction between nodes according to their structure.
\end{itemize}

\subsection{Baseline Methods}\label{sec:appe:base}
We adopted three mainstream graph data augmentation methods as our baselines, including DropEdge, DropNode, Subgraph  and Manifold-Mixup. The details of the baselines are listed as follows,
\begin{itemize}
    \item \textbf{DropEdge}\footnote{\url{https://github.com/DropEdge/DropEdge}}~\citep{rong2020dropedge}. DropEdge randomly removes a
certain ratios of edges from the input graph at each training epoch, which can  prevent over-fitting and alleviate over-smoothing.
    \item \textbf{DropNode}\footnote{\url{https://github.com/Shen-Lab/GraphCL}}~\citep{you2020graph}. DropNode randomly remove certain portion of nodes as well as their connections, which under a underlying assumption that missing part of nodes will note affect the semantic meaning of original graph.
    \item \textbf{Subgraph}\footnote{\url{https://github.com/Shen-Lab/GraphCL}}~\citep{you2020graph,wang2020graphcrop}. Subgraph method samples a subgraph from the original graph using random walk The generated graph will keep part of the the semantic meaning of original graphs.
    \item \textbf{M-Manifold}\footnote{\url{https://github.com/vanoracai/MixupForGraph}}~\citep{wang2021mixup} Manifold-Mixup conducts Mixup operation for graph classification in the embedding space, which interpolates graph-level embedding after the READOUT function.
\end{itemize}

\subsection{Experimental Setting of Robustness}\label{sec:appe:roub}
The graph neural network adopted in this experiment is GCN, the architecture of which is as above. For label corruption, we randomly corrupt the graph labels with different corruption ratio $10\%, 20\%, 30\%, 40\%$. For topology corruption, we we randomly remove/add edges with different corruption ratio $10\%, 20\%, 30\%, 40\%$. The dataset for topology corruption is REDDIT-BINARY.

\section{Additional Experiments}\label{sec:appe:exp}
In this appendix, we conduct additional experiments to further investigate the proposed method.

% \vspace{-5pt}
\begin{figure}[!tp]
      \centering
      \includegraphics[width=1.0\textwidth]{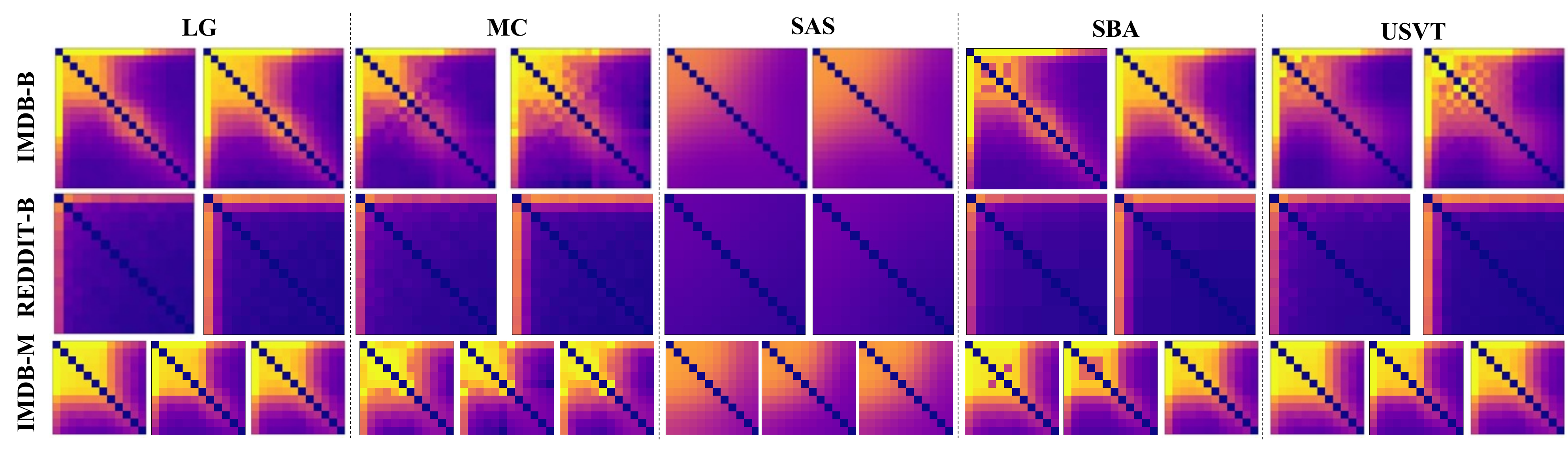}
      \vspace{-18pt}
      \caption{The estimated graphon on various dataset with different graphon estimation methods. }\label{fig:gon_ge1}
%   \vspace{-10pt}
\end{figure}

\subsection{Visualization of Graphons on More Real-world Dataset}\label{sec:appe:vis}

$\mathcal{G}$-Mixup explores five graphon estimation methods, including sorting-and-smoothing (SAS) method~\citep{chan2014consistent}, stochastic block approximation (SBA)~\citep{airoldi2013stochastic}, ``largest gap'' (LG)~\citep{channarond2012classification}, matrix completion (MC)~\citep{keshavan2010matrix} and the universal singular value thresholding (USVT)~\citep{chatterjee2015matrix}. We present the estimated graphon by $LG$ in \cref{fig:gon}. Here we present more visualization of graphons on IMDB-BINARY, REDDIT-BINARY and IMDB-MULTI dataset. An obvious observation is that graphons of different classes of graphs are different. This observation further validates the divergence of graphon between different classes of graphs.

\begin{table}[!tp]
\centering
\setlength\tabcolsep{4pt}
\caption{Performance comparisons of $\mathcal{G}$-Mixup with GMT on different dataset. The metric is the classification accuracy and its standard deviation. The best performance is in \textbf{boldface}.}\label{tab:gmt}
\begin{tabular}{llcccccccccc}
    \toprule
    \multicolumn{1}{l}{Backbone} &
    \multicolumn{1}{l}{Method} &
    \multicolumn{1}{c}{D\&D} &
    \multicolumn{1}{c}{MUTAG} & 
    \multicolumn{1}{c}{PROTEINS} & 
    \multicolumn{1}{c}{IMDB-B} &
    \multicolumn{1}{c}{IMDB-M} \\

%  \midrule
 \midrule
 \multirow{1}{*}{GMT}       &vanilla                  &\ms{78.29}{5.77}  &\ms{82.77}{6.30}  &\ms{74.59}{5.29}  &\ms{73.60}{3.87}  &\ms{50.73}{3.03}    \\
                            &w/ Dropedge              &\ms{78.37}{4.17}  &\ms{82.22}{8.88}  &\ms{74.32}{5.42}  &\ms{73.40}{3.85}  &\ms{50.73}{3.09}     \\
                            &w/ M-Mixup               &\ms{77.69}{3.81}  &\ms{82.22}{10.48} &\ms{74.41}{3.97}  &\ms{73.70}{3.79}  &\ms{49.93}{3.49}  \\
                            &w/ $\mathcal{G}$-Mixup   &\bms{79.57}{3.69}  &\bms{84.44}{8.88}  &\bms{75.13}{5.06}  &\bms{74.70}{3.76}  &\bms{51.33}{3.52}    \\
\bottomrule
\end{tabular}
\end{table}

\subsection{Experiment on More Graph Neural Networks Pooling Method (GMT)}\label{sec:appe:gmt}

To further validate the effectiveness of $\mathcal{G}$-Mixup on more graph neural networks, we experiment with GMT~\citep{baek2020accurate}, a modern pooling method.  To reproduce GMT results, we the released code and the recommended hyperparameters for their used datasets (D\&D, MUTAG, PROTEINS, IMDB-B, IMDB-M) in their paper. The results are presented in \cref{tab:gmt}. \Circled{\footnotesize 9}~\textbf{$\mathcal{G}$-Mixup can significantly improve the performance of GMT.}  \cref{tab:gmt} shows that $\mathcal{G}$-Mixup outperform all the baselines on all datasets. Overall, $\mathcal{G}$-Mixup outperform vanilla, Dropedge, ManifoldMixup by 1.44\%, 1.28\%, 2.01\%, respectively. This indicates the superiority of $\mathcal{G}$-Mixup for graph classification task.

\subsection{Experiment on Molecular Property Prediction}\label{sec:appe:ex_ogb}
We experiment on molecular property prediction task~\citep{hu2020open}, including ogbg-molhiv, ogbg-molbace, ogbg-molbbbp. In these dataset, each graph represents a molecule, where nodes are atoms, and edges are chemical bonds. We adopte official reference graph neural network backbones (gcn, gcn-vitual, gin, gin-vitual)~\footnote{\url{https://github.com/snap-stanford/ogb/tree/master/examples/graphproppred/mol}} as our backbones, and we generate the edge attributes randomly for synthetic graphs. The results are presented in \cref{tab:ogb}. \Circled{\footnotesize 10}~\textbf{$\mathcal{G}$-Mixup can improve the performance of GNNs on molecular property prediction task with the experimental setting for a fair comparison.} \cref{tab:ogb} shows that $\mathcal{G}$-Mixup gains $9$ best performances among $12$ reported AUCs.

\begin{table}[!tp]
    \centering
    \caption[Caption]{Performance comparisons of $\mathcal{G}$-Mixup on molecular property prediction task. The metric is AUROC~\footnotemark{} and its standard deviation. The best performance is in \textbf{boldface}. }\label{tab:ogb}
    % \scriptsize
    % \vspace{-10pt}
    \begin{tabular}{llcccc}
    \toprule
    \textbf{Backbones} &\textbf{Mehtods} &\textbf{ogbg-molhiv} &\textbf{ogbg-molbbbp} &\textbf{ogbg-molbace} \\\cmidrule{1-5}
    GCN         &vanilla 		                &\ms{76.24}{0.98}   &\ms{68.05}{1.52}   &\ms{80.36}{1.56} \\
                &w/ Dropedge 	                &\ms{75.93}{0.76}   &\ms{68.02}{0.95}   &\ms{80.22}{1.59} \\
                &w/ ManifoldMixup               &\ms{76.24}{1.40}   &\ms{68.36}{2.05}   &\ms{80.46}{2.05} \\
                &w/ $\mathcal{G}$-Mixup         &\bms{76.29}{0.80}  &\bms{69.51}{1.20} &\bms{80.73}{2.06} \\
    \midrule        
    GCN-virtual &vanilla                        &\ms{75.62}{1.65}   &\ms{65.13}{1.11}   &\bms{74.49}{3.04} \\
                &w/ Dropedge                    &\ms{74.64}{1.32}   &\ms{66.46}{1.61}   &\ms{69.75}{3.47} \\
                &w/ ManifoldMixup               &\ms{74.04}{2.06}   &\ms{65.51}{1.74}   &\ms{73.10}{4.97} \\
                &w/ $\mathcal{G}$-Mixup         &\bms{76.56}{0.80}  &\bms{70.05}{1.78}  &\ms{73.55}{4.79} \\
    \midrule        
    GIN         &vanilla                        &\ms{77.08}{1.96}   &\ms{68.42}{2.31}   &\ms{75.91}{1.01} \\
                &w/ Dropedge                    &\ms{75.77}{1.75}   &\ms{66.16}{2.96}   &\ms{70.50}{6.24} \\
                &w/ ManifoldMixup               &\ms{75.73}{1.25}   &\ms{68.15}{2.04}   &\ms{77.44}{4.13} \\
                &w/ $\mathcal{G}$-Mixup         &\bms{77.14}{0.45}  &\bms{70.17}{1.03}  &\bms{77.79}{3.34} \\
    \midrule        
    GIN-virtual &vanilla                        &\bms{77.52}{1.56}   &\ms{67.10}{2.10}   &\bms{74.19}{4.99} \\
                &w/ Dropedge                    &\ms{76.83}{1.11}   &\ms{68.87}{1.17}   &\ms{72.20}{3.37} \\
                &w/ ManifoldMixup               &\ms{76.51}{2.22}   &\ms{68.04}{2.87}   &\ms{74.17}{1.38} \\
                &w/ $\mathcal{G}$-Mixup         &\ms{77.09}{1.07}   &\bms{69.18}{0.87}   &\ms{73.53}{3.98} \\
    \bottomrule
    \end{tabular}
\end{table}
\footnotetext{The metric is Area Under Receiver Operating Characteristic.}

\subsection{Sensitivity Analysis to Mixup Ratio $\lambda$}\label{sec:appe:sensit}
To further investigate the performance of $\mathcal{G}$-Mixup, we provide experimental results of $\mathcal{G}$-Mixup to analyse the sensitivity to hyperparameter mixup ratio $\lambda$. Specifically, we use the different mixing ratio $\lambda$ in ${W}_{\mathcal{I}} = \lambda {W}_{\mathcal{G}} + (1-\lambda) {W}_{\mathcal{H}},\lambda \mathbf{ y }_{ \mathcal{G} } + (1-\lambda) \mathbf{y}_{\mathcal{H}}$ on molecular property prediction task (i.e., ogbg-molbbbp, ogbg-molbace). The p-value\footnote{A p-value less than 0.05 (typically $\le$ 0.05) is statistically significant.} is computed with the best performance compared to the Vanilla GCN (last column in \cref{tab:molbbbp_lam} and \cref{tab:molbbbp_lam}). We can observed that $\mathcal{G}$-Mixup significantly improves graph neural networks' performance while we tune the hyperparameter of $\mathcal{G}$-Mixup.

\begin{table}[!tp]
\centering
\caption{The sensitivity of $\mathcal{G}$-Mixup to Mixup Ratio $\lambda$ on ogbg-molbbbp dataset. The p-value is $0.00515,0.0994,0.0109,0.0471$ , indicating the $3$ improvements are statistically significant~($p< 0.05$).}
\setlength\tabcolsep{2pt}
\fontsize{9}{8}\selectfont
\label{tab:molbbbp_lam}
\begin{tabular}{lcccccccccc|c}
\toprule
\textbf{ $\lambda$ }  &\textbf{ 0.05 }     &\textbf{ 0.10 } &\textbf{ 0.15 } &\textbf{ 0.20 } &\textbf{ 0.25 }     &\textbf{ 0.30 } &\textbf{ 0.35 } &\textbf{ 0.40 } &\textbf{ 0.45 } &\textbf{ 0.50 }  &\textbf{ Vanilla } \\
\midrule
GCN         &\ms{68.23}{0.75}          &\ms{68.23}{1.81}      &\ms{68.45}{0.84}   &\ms{67.54}{3.09}      &\bms{69.51}{1.20} &\ms{67.79}{0.82}      &\ms{67.60}{1.31}      &\ms{69.48}{2.62}      &\ms{67.86}{1.02}      &\ms{68.78}{2.61}  &\ms{68.05}{1.52}\\
GCN-virtual &\ms{68.57}{2.61}          &\ms{68.81}{1.57}      &\ms{67.20}{1.30}   &\ms{68.64}{2.09}      &\bms{70.05}{1.78} &\ms{68.77}{2.31}      &\ms{69.11}{1.12}      &\ms{68.82}{0.98}      &\ms{69.07}{1.48}      &\ms{68.37}{0.95}  &\ms{65.13}{1.11}\\
GIN         &\ms{68.20}{1.04}          &\ms{69.37}{1.38}     &\ms{69.28}{1.24}   &\ms{68.89}{2.70}       &\ms{70.17}{1.03} &\ms{66.95}{0.92}      &\ms{69.86}{1.05}      &\ms{70.01}{1.14}      &\ms{68.65}{1.03}      &\ms{69.73}{1.32}   &\ms{68.42}{2.31}\\
GIN-virtual &\ms{70.58}{1.55} &\ms{69.44}{1.88}      &\ms{70.02}{1.68}   &\ms{69.77}{0.88}      &\ms{69.18}{0.87}          &\ms{68.17}{1.67}      &\ms{68.62}{1.15}      &\ms{69.16}{1.87}      &\ms{70.15}{1.32}      &\ms{68.66}{0.68}  &\ms{67.10}{2.10}\\
\bottomrule
\end{tabular}
\end{table}

\begin{table}[!tp]
\centering
\caption{The sensitivity of $\mathcal{G}$-Mixup to Mixup Ratio $\lambda$ on ogbg-molbace dataset. The p-value is $0.0227,0.0375,0.0401,0.0427$, indicating the $4$ improvements are statistically significant~($p < 0.05$).}\label{tab:molbace_lam}
\setlength\tabcolsep{2pt}
\fontsize{9}{8}\selectfont  

\begin{tabular}{lcccccccccc|c}
\toprule
\textbf{ $\lambda$ }  &\textbf{ 0.05 } &\textbf{ 0.10 } &\textbf{ 0.15}       &\textbf{ 0.20 } &\textbf{ 0.25 } &\textbf{ 0.30 } &\textbf{ 0.35 }     &\textbf{ 0.40 } &\textbf{ 0.45 } &\textbf{ 0.50 }  &\textbf{ Vanilla } \\
\midrule
GCN         &\ms{77.41}{2.24}      &\ms{77.33}{2.10}      &\bms{80.73}{2.06}    &\ms{78.42}{2.25}      &\ms{77.98}{2.03}      &\ms{79.25}{1.64}      &\ms{75.80}{4.31}          &\ms{78.40}{1.88}      &\ms{79.54}{1.25}        &\ms{77.90}{2.67}   &\ms{80.36}{1.56} \\
GCN-virtual &\ms{75.64}{4.03}      &\ms{76.80}{1.74}      &\ms{73.55}{4.79}     &\ms{76.46}{1.05}      &\ms{73.97}{4.11}      &\ms{76.55}{2.28}      &\ms{75.91}{2.73}          &\ms{77.99}{2.59}      &\bms{78.34}{1.10}       &\ms{72.84}{5.52}   &\ms{74.49}{3.04} \\
GIN         &\ms{76.44}{2.19}      &\ms{75.55}{4.05}      &\bms{77.79}{3.34}    &\ms{75.20}{2.91}      &\ms{74.79}{2.64}      &\ms{76.27}{4.61}      &\ms{73.02}{3.68}          &\ms{76.29}{3.55}      &\ms{75.77}{2.30}        &\ms{74.12}{4.12}   &\ms{75.91}{1.01} \\
GIN-virtual &\ms{74.51}{4.91}      &\ms{74.07}{2.76}      &\ms{73.53}{3.98}     &\ms{78.85}{1.98}      &\ms{77.15}{2.44}      &\ms{76.85}{3.42}      &\bms{79.69}{1.37}         &\ms{75.13}{5.46}      &\ms{77.04}{1.37}        &\ms{78.63}{2.04}   &\ms{74.19}{4.99}\\
\bottomrule
\end{tabular}
\end{table}

\end{document}